\documentclass[11pt]{article}

\usepackage[final]{acl}
\usepackage{longtable}
\usepackage{times}
\usepackage{latexsym}
\usepackage[T1]{fontenc}

\usepackage[utf8]{inputenc}

\usepackage{microtype}

\usepackage{inconsolata}

\usepackage{graphicx}
\usepackage{enumitem}
\usepackage{booktabs}
\usepackage{multirow}
\usepackage{amsmath}

\usepackage{algorithm}
\usepackage{algpseudocode}
\usepackage{array}

\usepackage[most]{tcolorbox}
\newtcolorbox[list inside=prompt,auto counter]{prompt}[1][]{
    colbacktitle=black!60,
    coltitle=white,
    fontupper=\footnotesize,
    boxsep=5pt,
    left=0pt,
    right=0pt,
    top=0pt,
    bottom=0pt,
    boxrule=1pt,
    #1,
}
\newcommand{\MIND}{\textsc{Mind}}

\title{\MIND: Towards Immersive Psychological Healing with \\Multi-Agent Inner Dialogue}

\author{
 \textbf{Yujia Chen\textsuperscript{1}\footnotemark[1]}, 
 \textbf{Changsong Li\textsuperscript{1}\footnotemark[1]}, 
 \textbf{Yiming Wang\textsuperscript{2}}, 
 \textbf{Tianjie Ju\textsuperscript{2,5}}, 
 \textbf{Qingqing Xiao\textsuperscript{3,4}}, \\
 \textbf{Nan Zhang\textsuperscript{1}}, 
 \textbf{Zifan Kong\textsuperscript{1}}, 
 \textbf{Peng Wang\textsuperscript{1}}, 
 \textbf{Binyu Yan\textsuperscript{1}\footnotemark[2]} \\
 \\
 \textsuperscript{1}Sichuan University  
 ~~\textsuperscript{2}Shanghai Jiao Tong University  
 \\  
 \textsuperscript{3}Mental Health Center, West China Hospital, Sichuan University\\
 \textsuperscript{4}WestChina School of Nursing, Sichuan University  
 \textsuperscript{5}National University of Singapore  
 \\
   \texttt{\{yujiachen0058,changsongli0018,alsaceym\}@gmail.com} \quad \texttt{\{yanby\}@scu.edu.cn}
}

\begin{document}
\maketitle
\footnotetext[1]{The two authors contribute equally.}
\footnotetext[2]{Corresponding author.}
\begin{abstract}
Mental health issues are worsening in today's competitive society, such as depression and anxiety. 
Traditional healings like counseling and chatbots fail to engage effectively, they often provide generic responses lacking emotional depth. Although large language models (LLMs) have the potential to create more human-like interactions, they still struggle to capture subtle emotions. This requires LLMs to be equipped with human-like adaptability and warmth.
To fill this gap, we propose the {\bf \MIND}~ (\textbf{M}ulti-agent \textbf{IN}ner \textbf{D}ialogue), a novel paradigm that provides more immersive psychological healing environments.
Considering the strong generative and role-playing ability of LLM agents, we predefine an interactive healing framework and assign LLM agents different roles within the framework to engage in interactive inner dialogues with users, thereby providing an immersive healing experience.
We conduct extensive human experiments in various real-world healing dimensions, and find that \MIND ~provides a more user-friendly experience than traditional paradigms. This demonstrates that \MIND ~effectively leverages the significant potential of LLMs in psychological healing.

\end{abstract}

\section{Introduction}

\begin{figure}[t]
  \includegraphics[width=\columnwidth]{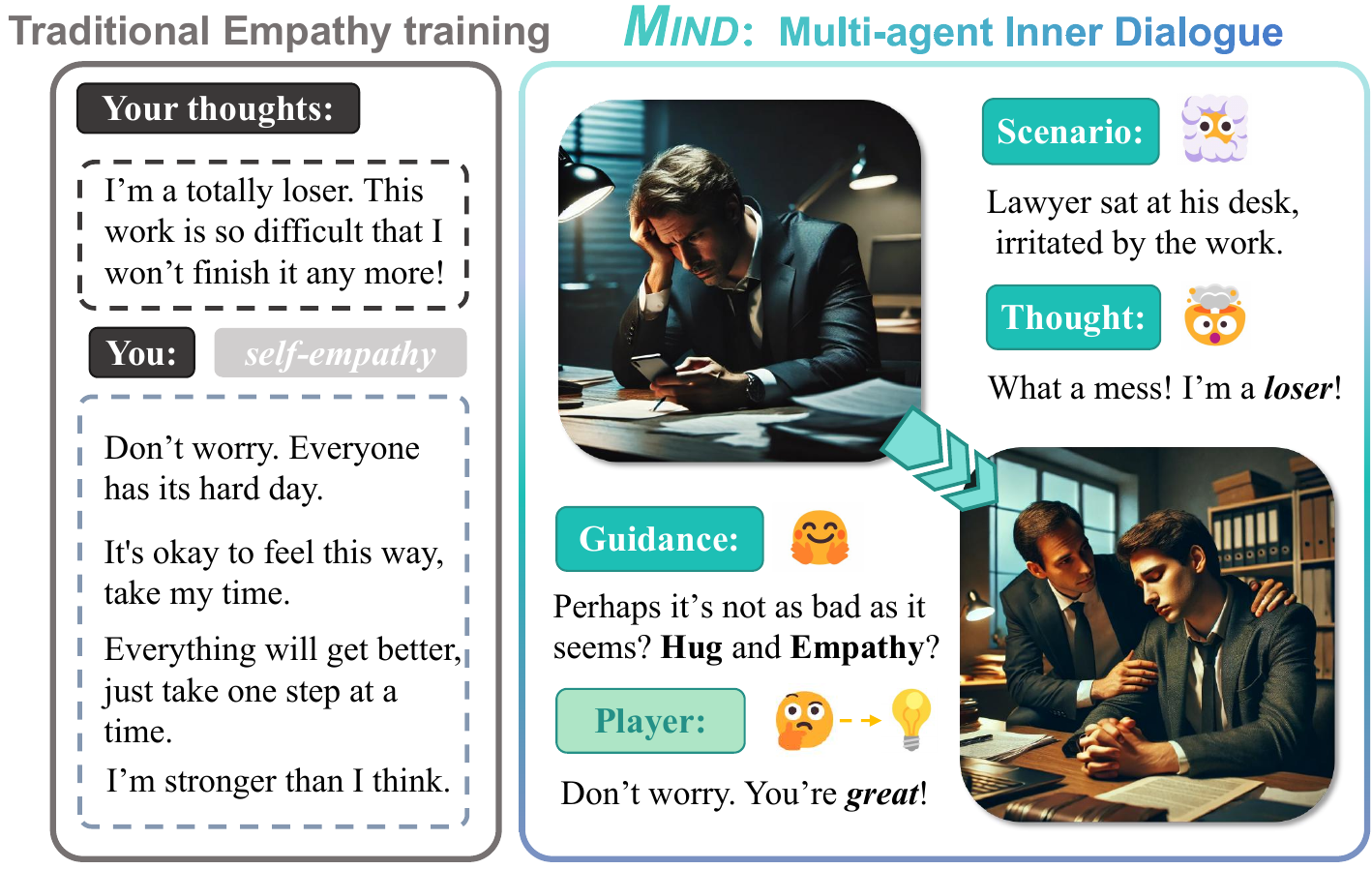}
  \caption{Examples of our \MIND ~paradigm with multi-agent inner dialogue compared to the traditional empathy training healing method.}
  \vspace{-0.1in}
  \label{fig:overview}
\end{figure}

Mental health issues are worsening in today's competitive society, with rising cases of disorders like depression \citep{moitra2023global}. This lead to a growing market for psychological healing. Traditional healing paradigms like Cognitive Behavioral Therapy \citep{beck1979cognitive} and Dialectical Behavior Therapy \citep{lynch2007dialectical} are widely used but rely on face-to-face interactions, making them time-consuming and costly \citep{duruz2003recherche} that limits large-scale accessibility.

Another healing paradigm, VR-based Empathy Training \citep{halim2023individualized,hidding2024single,dollinger2024virtual}, involves self-dialogue in virtual reality, where individuals alternate perspectives between comforting and being comforted through a virtual self-representation.
This process enhances self-empathy, thereby promoting self-compassion and reducing self-criticism.
However, current systems are limited by static scenarios and scripted interactions. 
The absence of counselor guidance and flexible feedback in these fixed frameworks limits emotional regulation and weakens the adaptability of therapy.


\begin{figure*}[t]
  \centering
  \includegraphics[width=\textwidth]{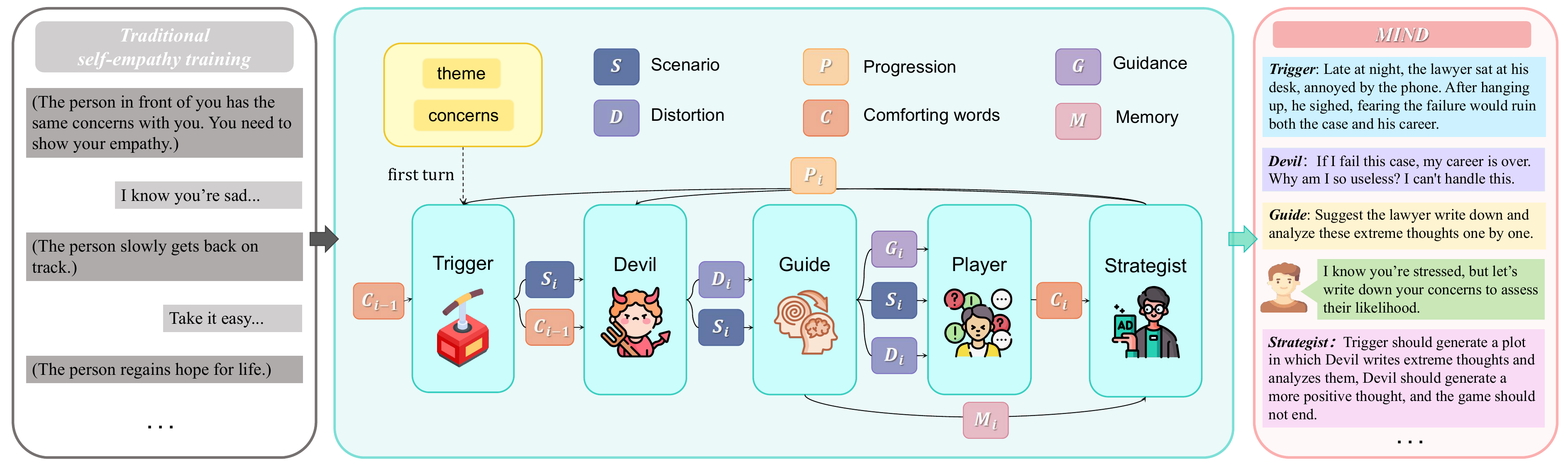}
  \caption{Overview of our \MIND ~paradigm: Trigger, Devil, Guide and Strategist interact with Player.}
    \label{fig:framework}
\end{figure*}

Recently, large language models (LLMs) have quickly advanced \citep{minaee2024largelanguagemodelssurvey,zhao2024surveylargelanguagemodels}, gaining strong abilities in generation \citep{X2025Learning}, reasoning \citep{huang-chang-2023-towards}, and role-playing \citep{wang-etal-2024-rolellm}. They also show great promise in mental health support \citep{hu2024psycollm,obradovich2024opportunities,bhatia2022cognitive}, offering new opportunities for psychological healing.
Despite these advancements, LLMs still face numerous challenges in the field of psychological healing. One major issue is {\it the lack of human empathy and the inability to form genuine therapeutic alliances}, which are crucial for effective treatment \citep{iftikhar2024therapy,guo2024large,obradovich2024opportunities,volkmer2024large}. LLMs often generate overly generic responses, failing to capture the subtle emotional nuances of patients \citep{sanu2024limitations}. These limitations highlight the need for a more sophisticated approach that blends LLMs' strengths with the warmth and flexibility of human interaction.

The emergence of multi-agent technology \citep{guo2024largelanguagemodelbased} offers potential solutions to these challenges. Multi-agent systems comprise specialized agents that collaborate and adapt to individual needs, ensuring a more immersive, interactive, dynamic healing experience \citep{guo2024multi,rocha2023applying}.
Each agent can focus on different aspects of psychological support, including emotional regulation, cognitive restructuring, and social interaction.
By utilizing the collective intelligence of multiple agents, they can provide a more comprehensive and effective experience.

Based on the above motivations, we propose {\bf M}ulti-agent {\bf IN}ner {\bf D}ialogue ({\bf \MIND}), a novel immersive and interactive psychological healing paradigm.
As illustrated in Figure \ref{fig:overview}, our approach is the first to introduce a multi-agent system into an empathy training paradigm, significantly enhancing the interaction between the user and their inner self through dynamic narrative scenarios.
Unlike traditional approaches that attempt to directly establish empathic alliances between therapists (or LLMs) and users, our framework emphasizes indirectly cultivating self-compassion in self-critical individuals. By assigning empathetic roles to LLM agents and directing empathy towards the user's own 
 ``inner self,'' our design enables users to engage in a simulated yet emotionally resonant ``self-to-self'' caring process. This mechanism is particularly beneficial for individuals who struggle to generate self-compassion through imagination or guided visualization.
We allocate four core roles to LLM agents (Trigger, Devil, Guide, and Strategist) each serving distinct reflective and emotional functions. 
This design helps foster internal empathy shifts and reveals cognitive patterns behind distress, promoting self-acceptance and psychological healing.

We conduct extensive experiments, including human evaluations, human experiments and ablation studies. The results demonstrate that:
\begin{itemize}
\item[$\bullet$]\textbf{\MIND~}outperforms traditional counseling, chatbots, and traditional empathy training methods,  achieving an average improvement of \textbf{13\%} across six psychological dimensions.
\item[$\bullet$]\textbf{\MIND~}demonstrates the highest positive emotional shift \textbf{(1.46)} and largest negative affect reduction \textbf{(-0.65)} among all dialogue systems, while also receiving the highest overall ratings from participants.
\item[$\bullet$] Ablation studies emphasize the significance of the memory mechanism, guide agent, and strategist agent, with an average performance drop of \textbf{42\%} when these components are removed.
\end{itemize}





\section{\MIND: Multi-agent Inner Dialogue}

\subsection{Overall Workflow}

The overall framework of our \MIND ~paradigm is shown in Figure \ref{fig:framework}, composed of four agents responsible for inner dialogue generation, in addition to an agent simulating patients with cognitive distortions. The subsequent section will commence with an overview of the workflow: the trigger, the devil, the guide, the strategist and the human simulated patient. Detailed prompt templates used by each agent are presented in Appendix \ref{sec:MIND_prompt}.

In this framework, $i$ stands for the $i$-th interaction. $S_i$ denotes the virtual scenarios. $D_i$ represents the distorted thoughts. $G_i$ refers to the professional psychological guidance. $C_i$ indicates the comforting words provided by the player. $M_i$ is the cumulative memory, which is a structured summary of previous scenarios, cognitive distortions. $P_i$ represents the storyline progression. Additionally, $W$ stands for the player's concerns, and $T$ denotes the overarching theme of the interaction.

\textbf{Step 0}: In the initial turn, the player articulates their current concern $W$ and selects a theme $T$, which together serve to anchor and guide the direction of the narrative. 

\textbf{Step 1}: The trigger $S_i$ is dynamically generated based on either the player's initial inputs $W$ and $T$ (in the first turn), or the reflective response $C_{i-1}$ and planning signal $P_{i-1}$ (in subsequent turns). $S_i$ is constructed to reflect the player's concerning scenes.

\textbf{Step 2}: The devil processes $S_i$, $C_{i-1}$ and $P_{i-1}$ to $D_i$, emulating maladaptive cognitive biases aligned with the player's mental state.

\textbf{Step 3}: The guide then integrates $S_i$ and $D_i$ to generate $G_i$, aimed at facilitating empathetic responses from the player. 

\textbf{Step 4}: Upon receiving $G_i$, the player engages in a reflective dialogue to provide $C_i$ and counter $D_i$, thereby advancing the therapeutic narrative.

\textbf{Step 5}: The strategist analyzes $M_{i-1}$ and $C_i$. This analysis produces $P_i$ that govern the generation of subsequent triggers ($S_{i+1}$) and the devil's adaptive cognitive evolution ($D_{i+1}$).




Through iterative cycles of scenario generation, cognitive reflection, and guided intervention, the framework progressively refines its alignment with the player's psychological profile. The entire algorithm is provided in the Algorithm \ref{sec:algorithm}.

\begin{algorithm}[h]
\caption{\MIND~ Paradigm}
\label{sec:algorithm}
\begin{algorithmic}[1]
\State \textbf{Input:} Player's concerns, Theme
\State \textbf{Output:} The player reaches a reconciliation with their own concerns.

\State \textbf{Initialize:} 
\State \quad Memory $M_0 \gets \emptyset$, iteration counter $i \gets 0$
\State \quad Generate initial scenario $S_0$ and initial distortion thoughts $D_0$ based on player's concerns and theme

\While{Player Engaged $\land$ $\neg$Therapeutic Goal Reached} \label{line:loop}
    \State \textbf{Step 1: Scenario Generation}
    \State \quad $S_i \gets \textsc{Scenario}(C_{i-1}, P_{i-1})$

    \State \textbf{Step 2: Distorted Thought Processing}
    \State \quad $D_i \gets \textsc{Distortions}(S_i, C_{i-1}, P_{i-1})$
    
    \State \textbf{Step 3: Psychological Guidance}
    \State \quad $G_i \gets \textsc{Guidance}(S_i, D_i)$
    
    \State \textbf{Step 4: Comforting Dialogue}
    \State \quad Present $S_i$, $D_i$, and $G_i$ to player
    \State \quad $C_i \gets \textsc{GetComfortingWords}()$
    
    \State \textbf{Step 5: Storyline Progression}
    \State \quad $P_i \gets \textsc{AnalyzeMemory}(M_{i-1}, C_i)$
    
    \State $i \gets i + 1$ \label{line:increment}
\EndWhile

\State \textbf{Output: }Enhanced therapeutic engagement and narrative continuity
\end{algorithmic}
\end{algorithm}


\subsection{Trigger: Scenario Generation}

The trigger generates artificial scenes within the interactive fiction game, drawing from the chosen theme and the player's concerns. It begins by creating an initial scene that reflects the player's psychological state and evolves the narrative based on previous interactions. The agent adapts the storyline according to the player's emotional context and worries, ensuring a coherent progression in the scene's development. Through this process, the trigger sets the stage for therapeutic reflection by crafting a dynamic and consistent narrative that mirrors the player's thoughts and psychological growth.

Let the first-round trigger agent be $\pi_{t_{0}}$ and non-first rounds trigger agent be $\pi_{t_{i}}$, the process can be formulated as:
\begin{equation}
    \begin{aligned}
        &S_{0}=\pi_{t_{0}}(W,T), \\
        &S_{i}=\pi_{t_{i}}(C_{i-1},P_{i-1};W,T) ~(i>0),
        \label{eq:trigger}
    \end{aligned}
\end{equation}

We adopt the chain-of-thought prompting technique \citep{wei2022chain} to enhance the quality of the trigger in scenario generation. Specifically, the trigger is instructed to generate a simulation scene based on the theme and the patient's concerns, while also explaining how to incorporate the scene history and the patient's thought processes to create a logical extension. 

\subsection{Devil: Cognitive Distortion Simulation}

The devil simulates the cognitive distortions that a patient might experience within the context of the scenario and it is aligned with the concept Simulated Patient(SP). It functions as the player's ``virtual embodiment'' representing an ``alternate self'' within the simulated environment.

Based on the simulated scenario provided by the trigger, the devil produces thoughts that align with common cognitive distortions, such as catastrophizing or emotional reasoning. These distortions are personalized to the player's specific context, offering an authentic simulation of how negative thinking can influence behavior and perceptions.

Let the first-round devil agent be $\pi_{d_{0}}$ and non-first rounds devil agent be $\pi_{d_{i}}$, the process can be formulated as:

\begin{equation}
    \begin{aligned}
        &D_{0}=\pi_{d_{0}}(W,S_{0}),\\
        &D_{i}=\pi_{d_{i}}(C_{i-1},P_{i-1},S_{i}) ~(i>0),
        \label{eq:devil}
    \end{aligned}
\end{equation}

To refine the simulation of the player's psychological state, we incorporate descriptions and definitions of five personality traits into the prompt design, aiming to create a more precise and personalized cognitive model. In the initial iteration, the devil agent generates responses solely based on the player's initial input and the scenario created by the trigger. However, in each subsequent iteration, the devil reacts to the player's comforting words, gradually weakening its cognitive distortions over time. This dynamic adjustment optimizes the player's interactive experience by allowing the devil's responses to evolve in alignment with the player's engagement and cognitive restructuring efforts.

\subsection{Guide: Cognitive Restructuring Guidance}

The guide aims to assist the player in recognizing, challenging, and reframing negative thought patterns through cognitive restructuring. The process begins with the guide identifying cognitive distortions in the player's thinking, which may have been amplified by the devil. The guide then offers alternative perspectives to counter these irrational beliefs and provides practical suggestions, such as taking a deep breath or writing down worries to evaluate their validity. The guide's goal is not to enforce immediate change, but to support gradual shifts in thinking, ensuring that each new perspective is integrated at the player's own pace.

Denote the guide agent as $\pi_{g}$. The process can be formulated as:

\begin{equation}
  \label{eq:guide}
  (G_{i},M_{i})=\pi_{g}(S_{i},D_{i})
\end{equation}

As the game progresses, the growing history becomes burdensome for the LLM to process efficiently. To mitigate this issue, a summarization mechanism is employed to maintain coherent narrative memory \citep{zhou2023recurrentgpt}. In our implementation, we use a well-designed prompt template (see Appendix \ref{sec:MIND_prompt} for details) to guide the model to extract key events, emotional states, and cognitive distortion patterns from the history of the interaction and save them to the memory unit. The system recursively compares new and old memories, merges redundant information, and retains core therapeutic cues such as ``from self-denial to initial reflection'' to ensure coherent and streamlined historical memory. By utilizing this summarization mechanism, the guide ensures that the player is not only challenged but also supported in a structured, manageable way, encouraging long-term emotional resilience and rational thinking. Ultimately, the guide helps transform the player from a passive recipient of distorted thoughts, as influenced by the devil, into an active participant in their own cognitive change, laying the foundation for healthier thought patterns and emotional well-being.

\begin{figure*}[ht]
  \centering
  \includegraphics[width=\textwidth]{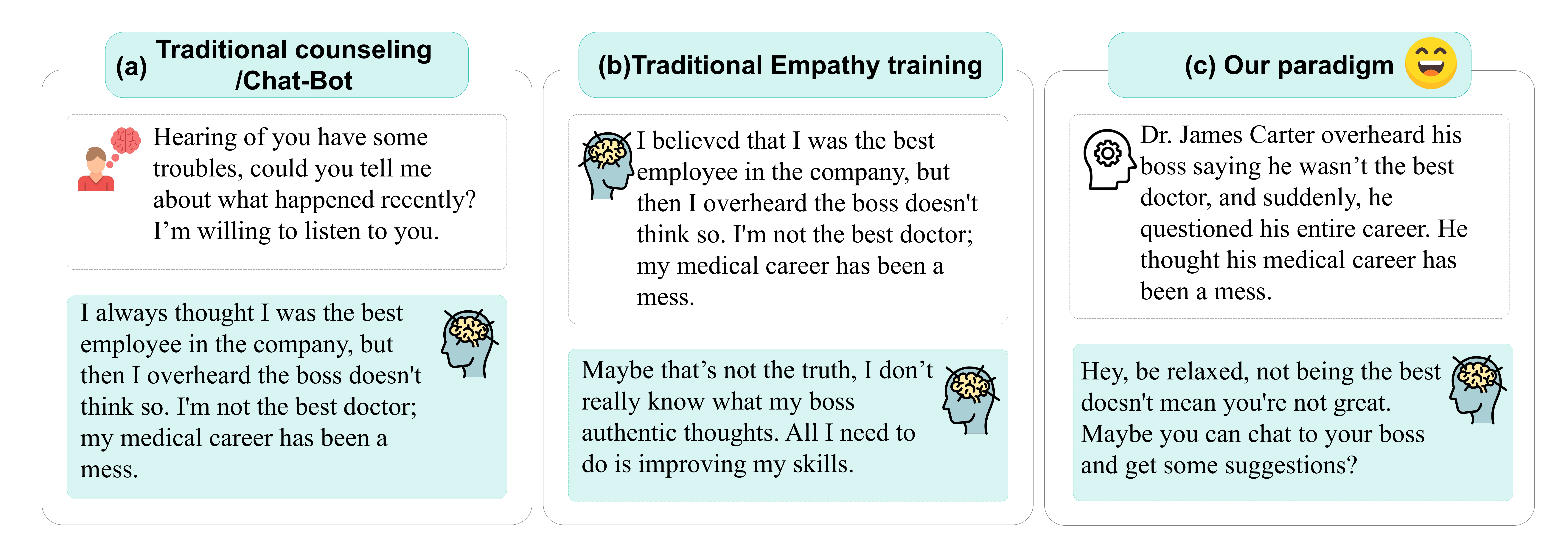}
  \caption{Comparison between three healing paradigms: Traditional counseling (or Chat-Bot), traditional empathy training and our paradigm. \MIND~ transfers a traditional healing environment into an artificial interactive scenario where players show empathy to their ``internal-self''.}
  \label{fig:paradigmcomparison}
\end{figure*}

\subsection{Strategist: Storyline Progression}

The strategist is responsible for planning the next stage of the narrative and determining the mental shifts of the antagonist based on previous events and the comfort provided by the player. The primary goal of the strategist is to ensure that the protagonist's cognitive distortions are gradually restructured through the unfolding of the story.

Denote the strategist agent as $\pi_{s}$. The process can be formulated as:

\begin{equation}
  \label{eq:strategist}
  P_{i}=\pi_{s}(M_{i},C_{i})
\end{equation}

In each iteration, the strategist carefully evaluates whether the devil's mindset has evolved. If the comforting words successfully address the devil's cognitive distortions, a shift in their thought process occurs, leading to a more balanced and realistic perspective on their circumstances. This change catalyzes the natural progression of the story, with the devil's actions and decisions reflecting a healthier mindset. Conversely, if no change takes place, the narrative remains consistent with the devil's previous emotional state, allowing the player's guidance to continue influencing their emotional transformation. The objective is to ensure that every story development is not only logically coherent but also aligns with the devil's cognitive journey toward self-awareness and emotional resilience.

\subsection{Human Simulated Patient: Empathy and Interaction}

To facilitate the automated operation and evaluation of our framework, and drawing upon the validated psychological characteristics and annotation capabilities of LLM, we employ LLMs to simulate players with cognitively distorted thinking and provide comforting words to the devil. Based on the guidance from the guide, the virtual scenario generated by the trigger, and the cognitive distortions produced by the devil, human simulated patient assumes the role of the Player, engaging in empathetic reassurance toward the devil. This process also incorporates the chain-of-thought (CoT) technique\citep{wei2022chain}, allowing for a structured and coherent response generation that aligns with the psychological progression of the player-agent interaction.

\section{Experiments}



\subsection{Setup}
\label{sec:content evaluation}

\paragraph{Scenario Setting.}
The real-life scenarios, thinking patterns, and cognitive distortion types of the Human Simulated Patient simulated by the LLM are derived from the C2D2 dataset \citep{wang2023c2d2}. This dataset is the first publicly available resource focused on cognitive distortion analysis, solving the problem of data scarcity in this field. The dataset covers eight major topics, including work issues, interpersonal issues, economic issues, random negative events, family issues, physical stress, and discrepancy between ideal and reality. All the experimental results in the body part were conducted in English, and the Chinese experimental results are presented in Appendix \ref{sec:Chinese_experiment}.

\paragraph{Baseline Paradigms.}
To evaluate the effectiveness of our \MIND ~paradigm, we compare it with traditional counseling methods (face-to-face dialogue and Q\&A) and the traditional empathy training paradigm \citep{halim2023individualized,hidding2024single,dollinger2024virtual}. Figure \ref{fig:paradigmcomparison} presents a comparison between these three paradigms, with the detailed implementation of baseline methods provided in Appendix \ref{sec:Baseline_methods}.

\paragraph{LLM Agents.}
We used several LLM agents including both open-source and closed-sourced models with varying parameter scales.
For closed-source models, we chose Gemini-2.0-flash \citep{gemini_url}, GPT-4o \citep{openai2024gpt4ocard}, GPT-3.5-Turbo \citep{ye2023comprehensivecapabilityanalysisgpt3}.
For open-source models, we chose Llama-3.1-8B-Instruct \citep{grattafiori2024llama3herdmodels}, Qwen2.5-72B-Instruct \citep{qwen2025qwen25technicalreport}, Qwen2.5-7B-Instruct \citep{qwen2025qwen25technicalreport} and Deepseek-R1 \citep{deepseekai2025deepseekr1incentivizingreasoningcapability}. We set the temperature of each model to 0.7.

\paragraph{Evaluation Metrics.}
The quality of the devil agent's responses is critical to this framework, as it reflects the player's internal ``cognitive distortions'' and must closely align with their ``inner voice''.
To ensure this, we first perform a preliminary SP role-playing evaluation in Section \ref{sec:sp role-playing assessment} to assess whether the model can accurately identify the type of cognitive distortion of the player and realistically express their thoughts.
We invited five mental health professionals, each of whom held 10 dialogue rounds with each model. They rated the responses using five evaluation metrics \citep{johri2025evaluation}, on a scale from 1 to 5.
Detailed evaluation metrics are shown in Appendix \ref{sec:SP metrics}.
Based on this experiment, we will select the best-performing model to conduct our main experiments.

In Section \ref{sec:main-results}, we conduct our main experiments to compare our \MIND~with other paradigms.
We evaluate three main aspects: {\it user experience}, {\it interaction quality}, and {\it emotional comfort}, with six different metrics \citep{hua2024large,kumaran2023scenecraft,jennett2008measuring,ryan2015narrative,nacke2011towards}. Metric details are shown in Appendix \ref{sec:Metrics_description}.
We recruited 7 mental health professionals with professional expertise in psychological therapy. For the different paradigms, the evaluators rated the content based on the six evaluation metrics, with a scoring range of 1 to 5. 

\begin{table}[t]
\begin{tabular}{p{3.5cm}p{0.3cm}p{0.3cm}p{0.3cm}p{0.3cm}p{0.3cm}}
    \toprule
    \textbf{Model Name} & \textbf{DS} & \textbf{CF} & \textbf{EE} & \textbf{PD} & \textbf{Acc} \\
    
    \midrule
    \multicolumn{6}{c}{\it Closed-Source Model} \\
    \midrule
    Gemini-2.0-flash & \textbf{\underline{4.8}} & 4.2 & \textbf{\underline{4.4}} & \textbf{\underline{4.6}} & 4.2 \\
    GPT-4o & \textbf{\underline{4.8}} & \textbf{\underline{4.4}} & 4.0 & 3.6 & \textbf{\underline{4.4}} \\
    GPT-3.5-Turbo & 4.2 & 4.2 & 3.6 & 3.4 & 3.4 \\
    
    \midrule
    \multicolumn{6}{c}{\it Open-Source Model} \\
    \midrule
    Qwen2.5-72B-Instruct & 3.2 & 2.8 & 3.0 & 2.6 & 3.0 \\
    Llama-3.1-8B-Instruct & 3.8 & 3.2 & 3.4 & 3.4 & 3.2 \\
    Qwen2.5-7B-Instruct & 3.2 & 2.8 & 3.0 & 2.8 & 3.0 \\
    Deepseek-R1 & 3.0 & 3.6 & 3.4 & 3.2 & 3.4 \\
    \bottomrule

\end{tabular}
\caption{SP role-playing results between different models. DS=Dialogue Stability, CF=Coherence \& Fluency, EE=Emotional Expression, PD=Personalization \& Diversity, Acc=Accuracy.}
\label{tab:spmodel}
\end{table}

\subsection{SP Role-playing Evaluation}
\label{sec:sp role-playing assessment}



We begin with a preliminary role-playing experiment to assess the performance of various models in the Simulated Patient (SP) role-playing task.
The results are presented in Table \ref{tab:spmodel}. Among these models, Gemini-2.0-flash performed best overall. While GPT-4o showed strength in some areas, it fell short in Emotional Expression and Personalization. Models such as GPT-3.5-Turbo, Llama-3.1-8B-Instruct, and Deepseek-R1 delivered weaker performance, especially in emotional and personalized responses. Qwen2.5 models ranked lowest, scoring below 3.2 across all dimensions, particularly in emotional expression and accuracy.
Based on these findings, {\bf we select Gemini-2.0-flash for our main experiments} due to its superior handling of the role-playing task and overall robustness.

\begin{figure}
  \centering
  \includegraphics[width=\columnwidth]{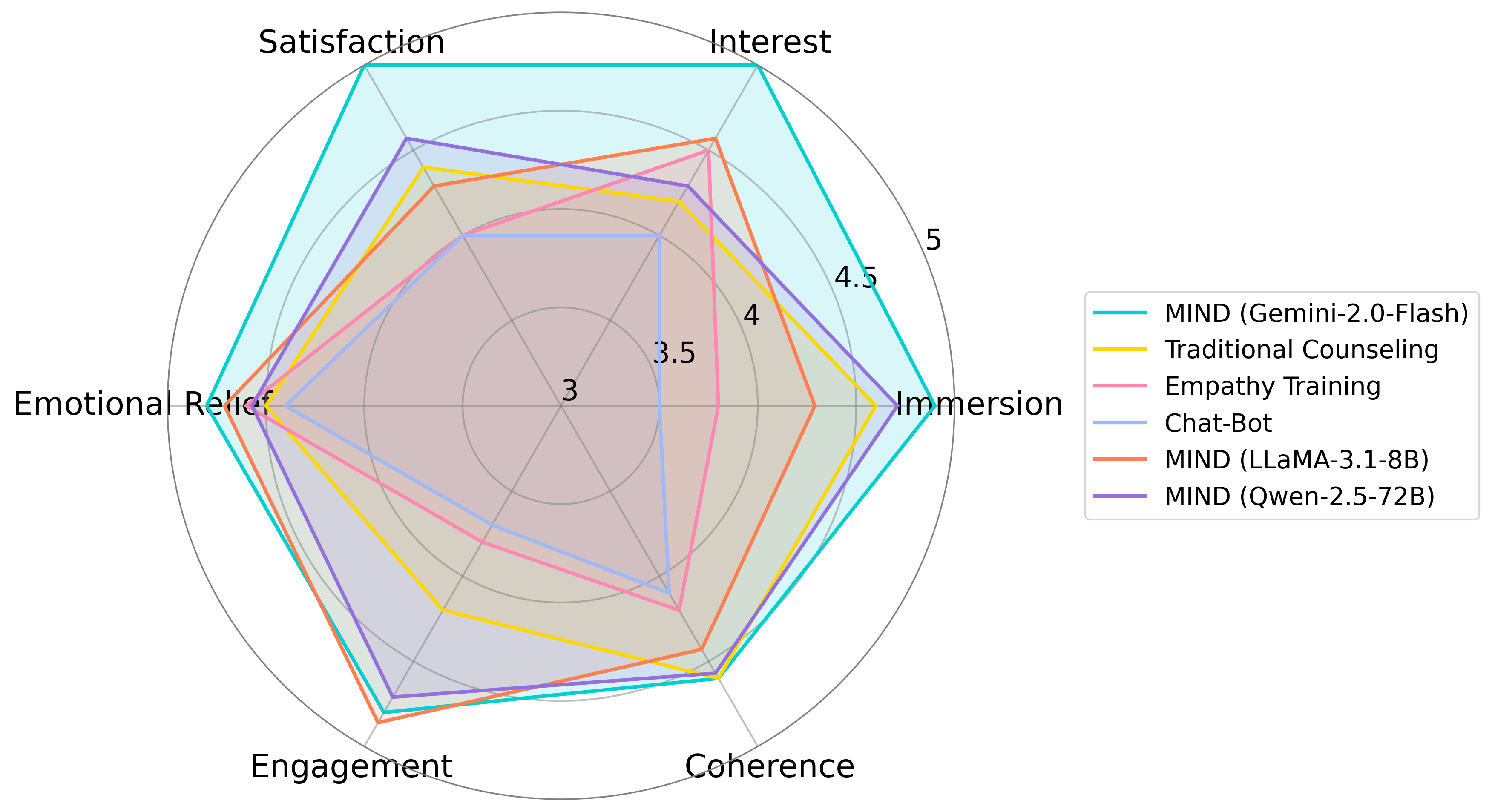}
  \caption{Comparisons among various healing methods through human evaluations. It is evident that our paradigm surpasses other paradigms in all aspects.}
  \label{fig:paradigmresult}
  \vspace{-0.1in}
\end{figure}

\begin{table*}
\centering
\begin{tabular}{cccccccc}
\hline
\textbf{Theme} & \textbf{IM}& \textbf{CO}& \textbf{EN}& \textbf{ER}& \textbf{SA} &\textbf{IN}\\
\hline
Work issues& 3.25& 3.00& 3.50& 3.25& 3.75& 3.75\\
Random negative events& 3.25& 3.50& 3.25& \textbf{3.75}& 3.50& 3.50\\
Interpersonal issues& \textbf{4.25}& 3.75& \textbf{4.25}& \textbf{3.75}& 4.25& \textbf{4.00}\\
Economic issues& 3.00& 4.00& 3.25& \textbf{3.75}& 3.75& 3.00\\
Family issues& 4.00& 3.75& 3.50& \textbf{3.75}& 3.50& 3.75\\
Physical stress& \textbf{4.25}& \textbf{4.25}& 3.75& \textbf{3.75}& 3.75& 3.75\\
Discrepancy between ideal and reality& \textbf{4.25}& 4.00& 3.75& \textbf{3.75}& \textbf{4.50}& 3.00\\
\hline
\end{tabular}
\caption{Content evaluation results between different themes. IM=Immersion, CO=Coherence, EN=Engagement, ER=Emotional Relief, SA=Satisfaction, IN=Interest. }
\label{tab:contenttheme}
\end{table*}

\begin{table*}[htbp]
\centering
\small
\begin{tabular}{lcccccccc}
\toprule
\textbf{Model} & \multicolumn{2}{c}{\textbf{EmoLLM}} & \multicolumn{2}{c}{\textbf{CACTUS}} & \multicolumn{2}{c}{\textbf{MIND}} & \multicolumn{2}{c}{\textbf{Control Group}} \\
\cmidrule(lr){2-3} \cmidrule(lr){4-5} \cmidrule(lr){6-7} \cmidrule(lr){8-9}
 & Positive & Negative & Positive & Negative & Positive & Negative & Positive & Negative \\
\midrule
\textbf{Average Fluctuation} & 0.36 & -0.11 & 1.35 & -0.52 & \textbf{1.46} & \textbf{-0.65} & -0.11 & 0.03 \\
\bottomrule
\end{tabular}
\caption{Comparison of average emotional fluctuation across different systems.}
\label{tab:emotional_fluctuation}
\end{table*}

\begin{table}[h]
\centering
\begin{tabular}{lccc}
\toprule
\textbf{Metrics} & \textbf{EmoLLM} & \textbf{CACTUS} & \textbf{MIND} \\
\midrule
IM & 2.5 & 3.5 & \textbf{5.0} \\
CO & 2.5 & \textbf{4.5} & \textbf{4.5} \\
EN & 2.0 & 4.0 & \textbf{4.5} \\
ER & 2.5 & 3.5 & \textbf{5.0} \\
SA & 2.0 & 4.0 & \textbf{5.0} \\
IN & 2.0 & 3.5 & \textbf{4.5} \\
\bottomrule
\end{tabular}
\caption{Client ratings for different systems across six evaluation dimensions. IM=Immersion, CO=Coherence, EN=Engagement, ER=Emotional Relief, SA=Satisfaction, IN=Interest.}
\label{tab:human_evaluation_scores}
\end{table}

\subsection{Main Results}
\label{sec:main-results}
The mean scores of each paradigm are shown in Figure \ref{fig:paradigmresult}. \MIND~demonstrated significant strengths in all six core assessment dimensions. Quantitative analysis showed that our paradigm performed particularly well on the dimensions of interest and satisfaction, reaching a perfect score of 5, compared to all the baseline methods of traditional counseling, traditional empathy training, and chat-bot. Notably, in terms of the engagement index, \MIND~achieved an absolute improvement of 17.1\% over the suboptimal method of traditional counseling, which reflects the increased motivation of the caller users that \MIND~can improve, so that they cooperate and participate in psychotherapy. On the dimensions of immersion, coherence and emotional relief, \MIND~also outperforms/equals the remaining three paradigms, which fully demonstrates that \textbf{\MIND~has the potential to advance psychological interventions by combining the scalability of LLMs with human-centered interaction design.}

While closed-source models such as Gemini-2.0-flash showed a slight advantage---likely attributable to their state-of-the-art capabilities in reasoning, cognition, and role simulation---the framework also enables open-source models to achieve effective therapeutic interactions, as is shown in Figure \ref{fig:paradigmresult}. We therefore emphasize that our methodology is not tied to any single model: it leverages the scalability of LLMs together with human-centered interaction design, making it broadly applicable across both closed-source and open-source systems for advancing psychological interventions.

\subsection{Human Experiment}
\label{sec:Human Experiment}
We recruited a total of 8 volunteers (3 males and 5 females) with similar age, educational background, and living conditions. To assess changes in clients' emotional states, we employed the Positive and Negative Affect Schedule (PANAS) questionnaire \citep{watson1988development}, which comprises 20 items covering 10 dimensions of positive and 10 dimensions of negative affect. Detailed experimental settings and the full PANAS questionnaire are provided in Table \ref{tab:panas}.

We calculated the average positive and negative emotional fluctuations of participants when interacting with three different systems: EmoLLM \citep{yang2024emollm}, CACTUS \citep{lee2024cactus}, \MIND, and a control group. The results are reported in Table \ref{tab:emotional_fluctuation}. In addition, participants rated each system across six subjective evaluation dimensions, with scores summarized in Table \ref{tab:human_evaluation_scores}.

As shown, \MIND~achieved the best overall performance, outperforming other systems both in terms of emotional improvement measured by PANAS and in subjective ratings across all six evaluation criteria.

\section{Analysis}
\label{sec:ablations}

\subsection{Thematic Scenarios Ablation}

This framework is applicable to a variety of thematic scenarios, including but not limited to work, family, and interpersonal issues. To analyze the differences in effectiveness across different themes within this framework, we independently generated five examples for each of the seven themes in the C2D2 dataset. Similarly, we invited evaluators with psychological therapy expertise to score these examples. As shown in Table \ref{tab:contenttheme}, the performance of different themes varies under our framework. Most themes perform well in ``Emotional Relief'' and ``Satisfaction'', indicating that the system can significantly alleviate users' emotions, fully exert its healing effects, and provide users with a positive experience. Immersion and Engagement are high, especially in themes like ``Physical stress'' and ``Interpersonal issues''. However, ``Work issues'' and ``Economic issues'' score lower in certain dimensions, which may require further optimization. 

\subsection{Agent Involvement Ablation}
Our framework consists of four agents: trigger, devil, guide, and strategist. To evaluate the effectiveness of \MIND's two core agents (i.e., the guide and strategist) as well as the memorization mechanism, we conducted several ablation experiments to assess their impact on user experience and demonstrate the importance of each component. Specifically, we randomly generated three examples for each ablation experiment. We recruited 4 clinical psychology researchers with professional expertise to evaluate six content evaluation metrics, as outlined in Table \ref{tab:content evaluation}.

\begin{figure}[t]
  \centering
  \includegraphics[width=0.9\columnwidth]{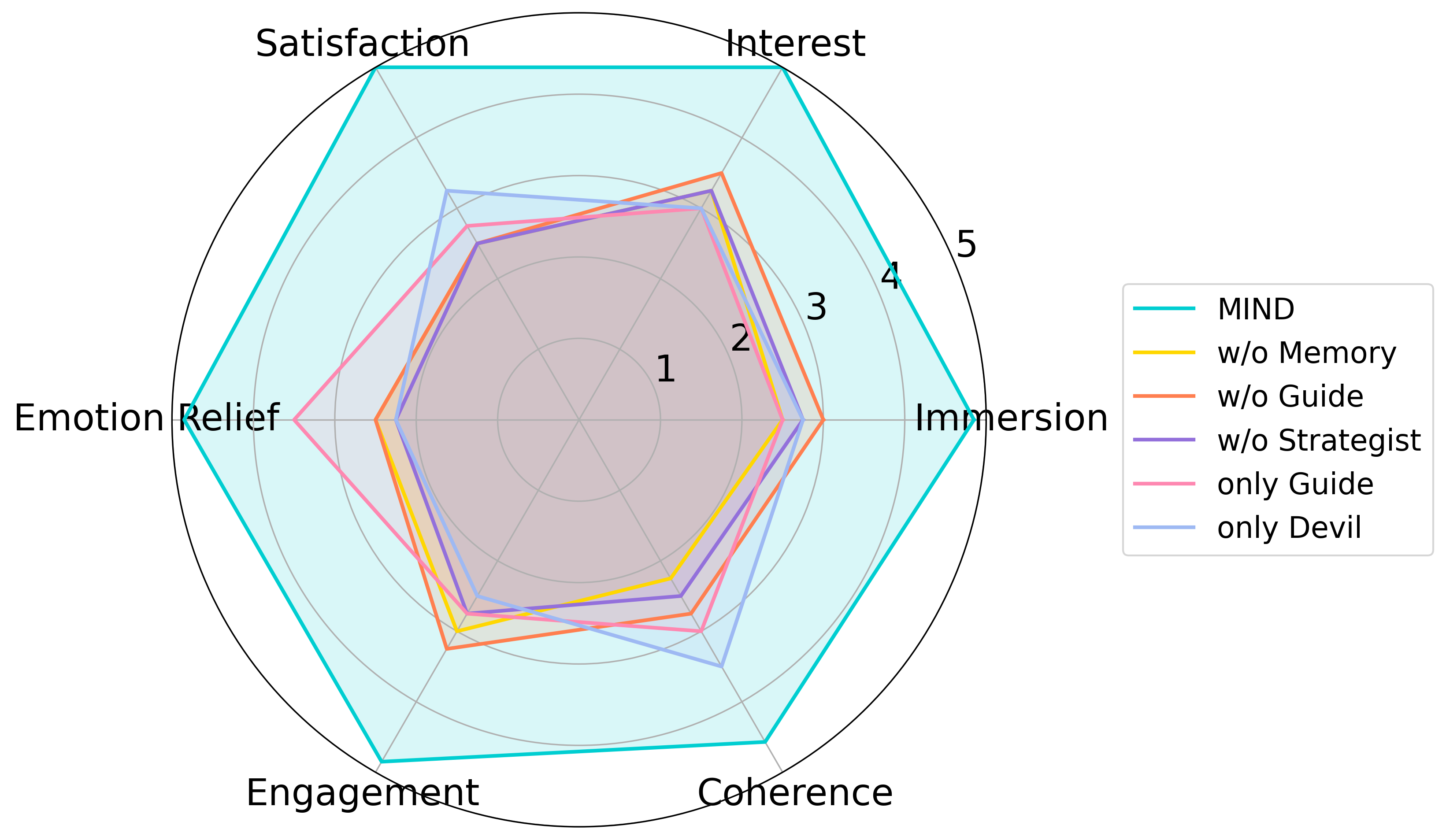}
  \caption{Ablations to assess the effectiveness
of \MIND~'s three agents (i.e., the guide, strategist and devil) and the memorization mechanism}
  \label{fig:agentablation}
\end{figure}

The experimental results are presented in Figure \ref{fig:agentablation}, which shows that {\bf each agent significantly contributes to the overall framework}. The removal of any agent or the memorization mechanism notably diminishes the quality of the generated content, underscoring the collective importance of all agents in the framework. Specifically, relying solely on the guide or the devil agent also leads to a decline in framework effectiveness, further underscoring the advantages of multi-agent collaboration.

\subsection{Strategist with Structured Facilitation Protocol}

One potential concern is the absence of an explicit facilitation protocol to manage stagnant conversations, which could risk inconsistent therapeutic progress in real-world applications. To examine this issue, we incorporated a structured facilitation protocol into the strategist agent's prompt design following \citep{seo2014personality}, with details provided in Appendix \ref{sec:MIND_prompt}. We then compared outcomes generated with and without this protocol using Gemini-2.0-flash. Across 70 samples (10 per theme for 7 themes), the proportion of failed cases-defined as instances where user thoughts were not fully shifted within 10 rounds-decreased only slightly, from 9\% to 6\%. This minimal improvement suggests that while the structured facilitation protocol provides an additional safeguard against conversational stagnation, its overall influence on performance is relatively limited.

\subsection{Case Study}

We present a case study in Appendix \ref{sec:case study}, featuring a four-round dialogue on the theme of ``work issues,'' with the concern: ``Despite studying hard, my grades remain poor, and effort seems useless in a talent-driven society.'' The case study shows how the devil agent gains confidence through the player's comforting words, while the player also develops greater self-compassion and reconciles with their own concerns.

\section{Related Work}

\subsection{LLM Agent}
An agent refers to an entity capable of perceiving its environment and taking action to achieve its goals. AI agents are increasingly seen as a promising direction toward achieving Artificial General Intelligence (AGI) \citep{durante2024agent}. Agents leverage the capabilities of Large Language Models (LLMs) to perform various tasks. In the construction of LLM agents, two of the most crucial aspects are (1) the architecture and (2) the method of acquiring capabilities. The architecture of LLM agents consists of four parts: Profile (primarily involving character background, written as prompts), Memory (including environmental and contextual information), Planning (allowing the agent to rationally execute according to a plan), and Action (transforming the agent's decisions into reasonable outputs)\citep{wang2024survey}. The method of acquiring capabilities is mainly divided into whether fine-tuning is performed. ReAct \citep{yao2022react} proposed a framework that combines reasoning and action, utilizing prompt engineering for task decomposition. Later, AutoGPT \citep{yang2023auto} introduced memory mechanisms and tool invocation capabilities, supporting multi-step task execution. HuggingGPT \citep{shen2024hugginggpt} coordinated multimodal models through LLMs, validating the potential of LLMs as the control hub. In multi-agent systems, early research borrowed from traditional multi-agent system architecture designs, proposing two mainstream frameworks: hierarchical (e.g., MetaGPT \citep{hong2023metagpt}) and decentralized (e.g., AutoGen \citep{wu2023autogen}). To enhance collaboration efficiency, researchers have explored various interaction paradigms, such as role-playing (CAMEL \citep{li2023camel} promotes task decomposition through predefined role divisions), debate negotiation (e.g., the debate decision-making framework MAD \citep{liang2024encouragingdivergentthinkinglarge}), and knowledge sharing (AgentVerse \citep{chen2023agentverse} uses dynamic memory banks to achieve experience transfer).

\subsection{LLM-assisted Psychology}
The powerful capabilities of LLMs in natural language processing and simulating interpersonal interactions have provided opportunities to assist in mental health. LLMs can play a role in various areas such as medical diagnosis, expansion of mental health resources, and therapy \citep{hua2024large}. In diagnosis, LLMs are widely used for screening and diagnosing mental health issues, including depression, anxiety, and post-traumatic stress disorder (PTSD). In mental health resource development, LLMs address the scarcity of mental health data by generating synthetic data (e.g., simulated counseling dialogues) or expanding existing clinical questionnaires. In psychological therapy, the application of LLMs offers new possibilities for improving mental health services. By increasing accessibility, providing personalized treatment plans, and reducing treatment costs, LLMs have the potential to enhance mental health care. SMILE utilizes ChatGPT to convert single-turn long conversations into multi-turn dialogues for the development of specialized dialogue systems for mental health support \citep{qiu2023smile}. SoulChat constructs the SoulChatCorpus dataset based on psychological consultation questions and answers, fine-tuning it to significantly enhance LLMs' abilities to provide empathy, listening, and comfort when offering emotional support \citep{chen2023soulchat}. MindChat is trained on one million high-quality multi-turn mental health conversation data to communicate in a more empathetic and guiding manner with users \citep{MindChat}. 

\section{Conclusion}

In this study, we propose \MIND ~paradigm, a novel paradigm for psychological healing. Our framework consists of four LLM agents: trigger, devil, guide, and strategist. Through iterative interactions between these agents and the player, the system comforts the player's ``inner self'' within a virtual scenario, thereby enhancing empathy and emotional resonance, reducing self-criticism, and fostering a stronger sense of self-identity. Experimental results validate the significant potential of this paradigm, demonstrating an improved user experience compared to both traditional psychological counseling models and the prototype of our framework. Our work provides a new perspective on gamified psychological healing and opens an innovative path for utilizing LLM agents in therapeutic applications. We hope this research offers a fresh outlook on the intersection of LLMs and psychological healing, encouraging the public to pay greater attention to and improve their mental health.

\section*{Ethics Statement}
The system used in this study is not intended to replace professional psychological treatment but rather to provide an effective option for clinical therapy. Before deployment, it is essential to ensure the presence of licensed professionals for supervision. Our evaluation method ensures the participation of mental health professionals and human experiment participants aged 18 and above. The human evaluators' ages range from 25 to 45 years, and their professions include one psychiatrist, two rehabilitation therapists, two psychotherapists, and two nurses. The human experiment participants' ages range from 18 to 21 years, and are all university students. Prior to the experiment, we provided the human evaluators with detailed experimental guidelines.

We have taken rigorous precautions to exclude individuals currently experiencing mental illness or those at risk of self-harm or suicidal tendencies. Our experiments are designed to avoid exposing participants to potentially harmful or misleading content. Participation in our evaluation experiment is entirely voluntary, and participants may withdraw at any time. We also ensured that a member of the research team was present throughout the process to guarantee its safety and effectiveness.

In our human study, we refrained from collecting any personally identifiable information, ensuring the anonymization of data before analysis. All research data were securely stored in a dedicated computing environment, accessible exclusively to trained research personnel.

\section*{Limitations}

While this study represents a significant step forward in shifting the paradigm of psychological healing, moving beyond the focus on training LLMs specifically for the psychological domain., it remains an initial attempt. To effectively implement this research into everyday psychological therapy, further extensive studies and clinical trials involving real mental health patients are necessary. Additionally, the framework's guide agent could benefit from being replaced with a more specialized therapeutic model, which could enhance the system's performance. Moreover, the framework used in this study is a simplified prototype. In the original theory , characters interact within a VR setting. There is significant potential for expanding this framework into more sophisticated formats, such as VR-based applications, to provide users with a more immersive and enriching therapeutic experience. Further exploration is required to address challenges related to the scalability of the system across various therapeutic scenarios and languages. Additionally, it remains unclear how the integration of this framework will scale in real-world settings with diverse patient populations, which presents another area for future research.

\section*{Acknowledgments}
We are appreciated to all patients and their legal guardians in the study. We thank Yiran Zhao (CTO, Shuyunxushi Technologies Co., Ltd., China) for research design and hardware support. We sincerely thank Xingmei Quan (Co-founder of Xinyan Community Psychological Service Development Center) for generously sharing her professional expertise in Cognitive Behavioral Therapy to support this study. We also acknowledge the platform support provided by the SCU Brain Health Research and Innovation Network research group (BRAINET). 

\bibliography{custom}
\clearpage
\appendix

\section{Baseline Methods}
\label{sec:Baseline_methods}

This section provides an overview of the baseline methods employed in our study. These methods serve as fundamental points of comparison, including two LLM-based baselines---\textbf{Chat-Bot} and \textbf{Traditional Empathy Training}---and one human-centered baseline, \textbf{Traditional Counseling}.

\textbf{Chat-Bot} refers to an LLM configured for psychological healing, simulating a virtual therapist that engages with users to provide emotional support. It identifies cognitive distortions during the conversation, offers comfort, and attempts cognitive restructuring. This baseline serves as a \emph{lower bound}, reflecting the performance of current single-agent psychological healing LLMs without the structured multi-agent framework proposed in our work.

\textbf{Traditional Empathy Training} employs role reversal in four phases to address cognitive distortions. In Phase 1, self-critical participants interact with a crying child avatar as an adult, demonstrating empathy. In Phase 2, some participants switch to the child avatar to receive comfort from their past selves, while others observe from a third-person perspective as a control. Phase 3 involves adapting to new perspectives: first-person participants embody the child avatar, while third-person participants observe without a virtual body. In Phase 4, participants re-experience empathy from the child's perspective, with real-time replays of the adult's gestures and voice.To better align with our current work, we simulated this process using LLMs. An agent, describing actions, demeanor, and emotions, played the role of the crying child. Participants provided verbal comfort and interacted with the agent, observing changes in the crying child. Once the interaction concluded (i.e., when the crying child stopped crying), the comforter assumed the child's perspective to review their comforting words and the child's responses, describing their psychological state. This approach, using agents, replicated the role reversal process typically conducted in Virtual Reality (VR), with prompts detailed in Appendix \ref{sec:MIND_prompt}. 

In our prompt design, the role of a little girl is adopted as the main character because this figure has been widely used in traditional VR-based empathy training. Moreover, it offers flexibility for personalization based on the patient's needs-for instance, it can be adapted to a little boy, an adult woman, or an adult man. This adaptability helps reduce the patient's psychological defensiveness while fostering empathy and a sense of care. As shown in Table~\ref{tab:deviation for different roles}, the variance in results across different character roles is relatively small, suggesting that the choice of role does not significantly affect the overall outcomes of the experiment.

\textbf{Traditional Counseling} refers to in-person or real-time professional psychotherapy, conducted by licensed human therapists. This condition serves as the gold standard baseline for human evaluation and is used for benchmarking the perceived effectiveness of our framework.

\section{SP Role-playing Assessment}
\label{sec:SP metrics}
We provide mental health professionals with the following statement to help them better comprehend tasks and assess models' all-round abilities.

\noindent
(1) \textbf{Dialogue Stability} 

Does the model consistently exhibit characteristics of cognitive distortion across all rounds of dialogue, rather than intermittently deviating from these traits? The simulated patient should maintain a stable mental state throughout the conversation, with consistency in the display of cognitive distortions. Furthermore, the content generated should reflect varying degrees of the same cognitive distortion type.

\noindent
(2) \textbf{Coherence \& Fluency}

Is the language coherent and fluent? Cognitive distortion patients may demonstrate features such as slowed speech, increased pauses, and disrupted speech patterns. The SP should replicate these linguistic tendencies, ensuring the language style aligns with the patient's condition and avoids inconsistencies.

\noindent
(3) \textbf{Emotional Expression}

Does the emotional content generated align with the emotional traits typical of cognitive distortion patients? The simulation should accurately reflect common emotional responses observed in these patients, such as persistent low mood, anhedonia, feelings of helplessness, and hopelessness.

\noindent
(4) \textbf{Personalization \& Diversity} 

In addition to core characteristics, does the model incorporate a wide range of individualized traits, such as how different personality traits, life experiences, and educational backgrounds influence the patient's expression and behavior? For example, introverted patients may exhibit more passive and reticent communication styles, while extroverted patients may display more outward and active engagement. The model should construct diverse cognitive profiles to ensure the simulated patient is both authentic and personalized by considering various influencing factors.

\noindent
(5) \textbf{Accuracy}

Is the identification of cognitive distortion types precise? This should be particularly evident in distinguishing the predominant distortion types when multiple cognitive distortions are present in the same interaction.

\onecolumn

\begin{table}[htp]
\centering
\begin{tabular}{@{}llcc@{}}
\toprule
\textbf{Metric} & \textbf{Character} & \textbf{Average} & \textbf{Standard Deviation} \\ \midrule
\multirow{5}{*}{Immersion} 
& little girl & 4.00 &  \\
& little boy & 3.75 &  \\
& woman & 3.25 & 0.53 \\
& man & 3.25 &  \\
& self in mirror image & 4.50 &  \\ \midrule
\multirow{5}{*}{Coherence} 
& little girl & 4.00 &  \\
& little boy & 3.25 &  \\
& woman & 4.00 & 0.31 \\
& man & 3.75 &  \\
& self in mirror image & 3.75 &  \\ \midrule
\multirow{4}{*}{Engagement} 
& little girl & 3.50 &  \\
& little boy & 3.25 &  \\
& woman & 3.50 & 0.47 \\
& man & 3.00 &  \\
& self in mirror image & 4.25 &  \\ \midrule
\multirow{4}{*}{Emotional Relief} 
& little girl & 3.75 &  \\
& little boy & 3.50 &  \\
& woman & 3.50 & 0.45 \\
& man & 3.00 &  \\
& self in mirror image & 4.25 &  \\ \midrule
\multirow{4}{*}{Satisfaction} 
& little girl & 4.00 &  \\
& little boy & 3.50 &  \\
& woman & 3.25 & \textbf{0.54} \\
& man & 3.25 &  \\
& self in mirror image & 4.50 &  \\ \midrule
\multirow{4}{*}{Interest} 
& little girl & 3.50 &  \\
& little boy & 3.00 &  \\
& woman & 4.00 & 0.50 \\
& man & 3.00 &  \\
& self in mirror image & 4.00 &  \\ \bottomrule
\end{tabular}
\caption{Average and Standard Deviation for Metrics Across Different Roles}
\label{tab:deviation for different roles}
\end{table}

\twocolumn

\twocolumn
\section{Human Experiment Details}
\label{sec:Human Experiment Details}
\subsection{Pipeline}

\noindent
\textbf{Step 1: Participant Recruitment and Screening }

We recruited a total of 8 volunteers, 3 males and 5 females, and similar in age, educational background, and living situation. There were 2 volunteers who did not participate in the model interaction and were only recruited to compare the likelihood of natural fluctuations in mood over time.
Participants were required to have worries that bothered them for 1 day to 1 week. We paid 50rmb per participant as a subsidy.

\noindent
\textbf{Step 2: Pre-test Evaluation} 

We measured clients' mood changes using the Positive and Negative Affect Scale (PANAS) questionnaire. The questionnaire contains 20 questions covering 10 positive and 10 negative emotion dimensions.
Prior to the start of the experiment, our coauthor mental health experts introduced the PANAS questionnaire and the scoring criteria of the six dimensions we proposed, and informed participants that they could terminate the experiment at any time, and that the experimental data would be kept completely confidential and anonymized, so that they should fill in the form as honestly and as naturally as possible, and give the feedback that most closely corresponded to their inner thoughts.

\noindent
\textbf{Step 3: Experimental Design and Model Assignment }

We randomly assigned the six participants in the experimental group to three systems (EmoLLM, CACTUS, \MIND), with each system corresponding to two clients.
The control group did not engage in any dialog and only filled out the questionnaire twice (30 minutes apart).

\noindent
\textbf{Step 4: Experimental implementation}

Participants entered a real counseling room and engaged in five rounds of text-based conversations with the assigned model via a computer. The content of the conversations was kept strictly confidential, and the model stopped recording as soon as the conversations were over.
Participants were asked to communicate about their ``short-term negative experiences'', such as academic stress, relationship problems, and so on.

\noindent
\textbf{Step 5: Post-test and data collection}

After the dialogues, participants completed the PANAS questionnaire again to compare the change in mood (e.g., whether the negative mood score decreased). All the emotion ratings are presented in Table \ref{tab:client_emotion_results}
And anonymous feedback on the system was collected. We collected the results as follows:

Client 5 stated that MIND is helpful for emotion channeling, and that it can give a positive suggestion to oneself by consoling others. Client 3 affirmed CACTUS's emotion channeling ability, but said that the response style is ``a little bit formatted and not very flexible'', and Client 1 is skeptical about EmoLLM, thinking that ``the content is empty, a lot of words, but there is no useful information, with low emotional value, unattractive''. Client 2 and Client 6 mentions the ``problem of long response time'', which may be a major constraint to the application of LLM in real-life counseling scenarios.

\onecolumn

\begin{table}[!h]
  \centering
  \footnotesize
  \resizebox{\textwidth}{!}{
  \begin{tabular}{@{}lp{11cm}@{}}
    \toprule
    \multicolumn{2}{l}{\textbf{Positive and Negative Affect Schedule (PANAS)}} \\
    \midrule
    \multicolumn{2}{l}{\textbf{I. Positive Affect}} \\
    \midrule
    1. & Interested \\
       & A. Very Rarely or Not at All  B. Very Little  C. Moderately  D. Quite a Bit  E. Very Much \\
    2. & Excited \\
       & A. Very Rarely or Not at All  B. Very Little  C. Moderately  D. Quite a Bit  E. Very Much \\
    3. & Strong \\
       & A. Very Rarely or Not at All  B. Very Little  C. Moderately  D. Quite a Bit  E. Very Much \\
    4. & Enthusiastic \\
       & A. Very Rarely or Not at All  B. Very Little  C. Moderately  D. Quite a Bit  E. Very Much \\
    5. & Proud \\
       & A. Very Rarely or Not at All  B. Very Little  C. Moderately  D. Quite a Bit  E. Very Much \\
    6. & Alert \\
       & A. Very Rarely or Not at All  B. Very Little  C. Moderately  D. Quite a Bit  E. Very Much \\
    7. & Inspired \\
       & A. Very Rarely or Not at All  B. Very Little  C. Moderately  D. Quite a Bit  E. Very Much \\
    8. & Determined \\
       & A. Very Rarely or Not at All  B. Very Little  C. Moderately  D. Quite a Bit  E. Very Much \\
    9. & Attentive \\
       & A. Very Rarely or Not at All  B. Very Little  C. Moderately  D. Quite a Bit  E. Very Much \\
    10. & Active \\
        & A. Very Rarely or Not at All  B. Very Little  C. Moderately  D. Quite a Bit  E. Very Much \\
    \midrule
    \multicolumn{2}{l}{\textbf{II. Negative Affect}} \\
    \midrule
    11. & Distressed \\
        & A. Very Rarely or Not at All  B. Very Little  C. Moderately  D. Quite a Bit  E. Very Much \\
    12. & Upset \\
        & A. Very Rarely or Not at All  B. Very Little  C. Moderately  D. Quite a Bit  E. Very Much \\
    13. & Guilty \\
        & A. Very Rarely or Not at All  B. Very Little  C. Moderately  D. Quite a Bit  E. Very Much \\
    14. & Scared \\
        & A. Very Rarely or Not at All  B. Very Little  C. Moderately  D. Quite a Bit  E. Very Much \\
    15. & Hostile \\
        & A. Very Rarely or Not at All  B. Very Little  C. Moderately  D. Quite a Bit  E. Very Much \\
    16. & Irritable \\
        & A. Very Rarely or Not at All  B. Very Little  C. Moderately  D. Quite a Bit  E. Very Much \\
    17. & Ashamed \\
        & A. Very Rarely or Not at All  B. Very Little  C. Moderately  D. Quite a Bit  E. Very Much \\
    18. & Nervous \\
        & A. Very Rarely or Not at All  B. Very Little  C. Moderately  D. Quite a Bit  E. Very Much \\
    19. & Jittery \\
        & A. Very Rarely or Not at All  B. Very Little  C. Moderately  D. Quite a Bit  E. Very Much \\
    20. & Afraid \\
        & A. Very Rarely or Not at All  B. Very Little  C. Moderately  D. Quite a Bit  E. Very Much \\
    \bottomrule
  \end{tabular}
  }
  \caption{The Questionnaire Measuring the Emotions of a Client based on PANAS}
  \label{tab:panas}
\end{table}

\begin{table}[htbp]
\centering
\resizebox{\textwidth}{!}{%
\begin{tabular}{@{}l*{24}{r}@{}}
\toprule
\multirow{2}{*}{Emotion} & \multicolumn{6}{c}{EmoLLM} & \multicolumn{6}{c}{CACTUS} & \multicolumn{6}{c}{MIND} & \multicolumn{6}{c}{Control Group} \\
\cmidrule(lr){2-7} \cmidrule(lr){8-13} \cmidrule(lr){14-19} \cmidrule(lr){20-25}
 & \multicolumn{3}{c}{client1} & \multicolumn{3}{c}{client2} & \multicolumn{3}{c}{client3} & \multicolumn{3}{c}{client4} & \multicolumn{3}{c}{client5} & \multicolumn{3}{c}{client6} & \multicolumn{3}{c}{client7} & \multicolumn{3}{c}{client8}\\
\cmidrule(lr){2-4} \cmidrule(lr){5-7} \cmidrule(lr){8-10} \cmidrule(lr){11-13} \cmidrule(lr){14-16} \cmidrule(lr){17-19} \cmidrule(lr){20-22} \cmidrule(lr){23-25}
 & b & a & $\delta$ & b & a & $\delta$ & b & a & $\delta$ & b & a & $\delta$ & b & a & $\delta$ & b & a & $\delta$ & b & a & $\delta$ & b & a & $\delta$\\
\midrule
Interested    & 1 & 2 & 1 & 1 & 3 & 2 & 1 & 4 & 3 & 2 & 4 & 2 & 1 & 4 & 3 & 2 & 4 & 2 & 2 & 1 & -1 & 1 & 1 & 0 \\
Excited       & 2 & 3 & 1 & 2 & 2 & 0 & 1 & 4 & 3 & 2 & 4 & 2 & 2 & 4 & 2 & 2 & 4 & 2 & 2 & 1 & -1 & 2 & 1 & -1 \\
Strong        & 2 & 4 & 2 & 2 & 3 & 1 & 1 & 4 & 3 & 3 & 4 & 1 & 1 & 5 & 4 & 2 & 5 & 3 & 2 & 1 & -1 & 2 & 2 & 0 \\
Enthusiastic  & 1 & 2 & 1 & 2 & 3 & 1 & 3 & 3 & 0 & 2 & 4 & 2 & 1 & 3 & 2 & 2 & 4 & 2 & 2 & 2 & 0 & 2 & 2 & 0 \\
Proud         & 2 & 3 & 1 & 2 & 2 & 0 & 2 & 4 & 2 & 2 & 5 & 3 & 2 & 4 & 2 & 2 & 5 & 3 & 1 & 1 & 0 & 1 & 1 & 0 \\
Alert         & 3 & 3 & 0 & 1 & 2 & 1 & 2 & 4 & 2 & 2 & 4 & 2 & 2 & 3 & 1 & 1 & 3 & 2 & 5 & 5 & 0 & 2 & 2 & 0 \\
Inspired      & 2 & 3 & 1 & 2 & 1 & -1 & 1 & 4 & 3 & 2 & 5 & 3 & 1 & 4 & 3 & 2 & 5 & 3 & 1 & 1 & 0 & 2 & 2 & 0 \\
Determined    & 2 & 3 & 1 & 2 & 1 & -1 & 1 & 4 & 3 & 2 & 4 & 2 & 1 & 4 & 3 & 2 & 5 & 3 & 1 & 1 & 0 & 2 & 2 & 0 \\
Attentive     & 3 & 4 & 1 & 2 & 3 & 1 & 1 & 4 & 3 & 2 & 4 & 2 & 2 & 5 & 3 & 2 & 4 & 2 & 1 & 2 & 1 & 2 & 2 & 0 \\
Active        & 2 & 3 & 1 & 2 & 2 & 0 & 1 & 4 & 3 & 3 & 4 & 1 & 2 & 4 & 2 & 2 & 4 & 2 & 1 & 1 & 0 & 2 & 1 & -1 \\
Distressed    & 3 & 4 & 1 & 4 & 2 & -2 & 4 & 2 & -2 & 4 & 1 & -3 & 4 & 2 & -2 & 4 & 2 & -2 & 4 & 4 & 0 & 4 & 4 & 0 \\
Upset         & 4 & 4 & 0 & 5 & 4 & -1 & 4 & 2 & -2 & 4 & 2 & -2 & 4 & 1 & -3 & 4 & 1 & -3 & 5 & 5 & 0 & 4 & 4 & 0 \\
Guilty        & 3 & 2 & -1 & 4 & 4 & 0 & 4 & 2 & -2 & 3 & 2 & -1 & 3 & 1 & -2 & 3 & 1 & -2 & 3 & 3 & 0 & 3 & 3 & 0 \\
Scared        & 4 & 4 & 0 & 4 & 3 & -1 & 5 & 1 & -4 & 4 & 2 & -2 & 4 & 1 & -3 & 4 & 1 & -3 & 3 & 3 & 0 & 3 & 3 & 0 \\
Hostile       & 3 & 2 & -1 & 1 & 2 & 1 & 5 & 2 & -3 & 4 & 2 & -2 & 4 & 1 & -3 & 3 & 1 & -2 & 3 & 4 & 1 & 4 & 4 & 0 \\
Irritable     & 3 & 3 & 0 & 1 & 3 & 2 & 5 & 2 & -3 & 4 & 1 & -3 & 4 & 2 & -2 & 4 & 1 & -3 & 4 & 3 & -1 & 3 & 4 & 1 \\
Ashamed       & 4 & 3 & -1 & 5 & 4 & -1 & 4 & 3 & -1 & 3 & 3 & 0 & 4 & 2 & -2 & 3 & 2 & -1 & 3 & 3 & 0 & 3 & 2 & -1 \\
Nervous       & 4 & 3 & -1 & 4 & 4 & 0 & 4 & 2 & -2 & 4 & 2 & -2 & 4 & 1 & -3 & 4 & 1 & -3 & 4 & 5 & 1 & 3 & 2 & -1 \\
Jittery       & 4 & 3 & -1 & 4 & 3 & -1 & 4 & 2 & -2 & 4 & 2 & -2 & 4 & 1 & -3 & 4 & 1 & -3 & 4 & 5 & 1 & 4 & 5 & 1 \\
Afraid        & 3 & 2 & -1 & 4 & 4 & 0 & 4 & 2 & -2 & 3 & 1 & -2 & 5 & 2 & -3 & 4 & 2 & -2 & 3 & 4 & 1 & 3 & 2 & -1 \\
\bottomrule
\end{tabular}%
}
\caption{Changes in PANAS Scores for Eight Clients Pre- and Post-Intervention. $b$ indicates scores before the intervention, $a$ represents scores after the intervention, and $\delta$ denotes the change calculated as post-intervention scores minus pre-intervention scores.}
\label{tab:client_emotion_results}
\end{table}
\newpage
\section{Chinese Experimental Results}
\label{sec:Chinese_experiment}
\begin{figure}[ht]
  \centering
  \includegraphics[width=0.4\columnwidth]{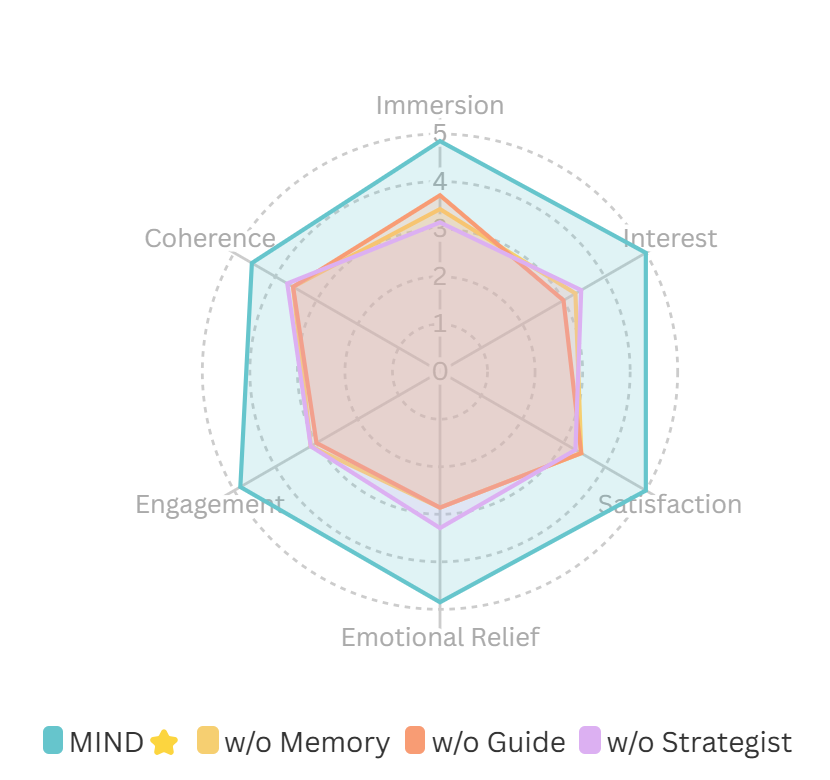}
  \caption{Ablations to assess the effectiveness
of \MIND~'s two agents (i.e., the guide and strategist) and the memorization mechanism}
  \label{fig:chineseagentablation}
\end{figure}

In this section, we present the results of the experiments conducted in Chinese, with the same experimental setup except for the language difference. 
Table \ref{tab:chinesecontenttheme} illustrates the results of the scene ablation experiment conducted in Chinese, indicating that our system exhibits stability across different scenarios. Figure \ref{fig:chineseagentablation} presents the results of the agent ablation experiment, indicating that the absence of the agent leads to a significant decline in outcomes regardless of the language used. This further demonstrates the rationality of our architecture.
\\
\begin{table*}[ht]
\centering
\renewcommand\arraystretch{1.1}
\begin{tabular}{cccccccc}
\hline
\textbf{Theme} & \textbf{IM}& \textbf{CO}& \textbf{EN}& \textbf{ER}& \textbf{SA} &\textbf{IN}\\
\hline
Work issues& 4.14& \textbf{4.71}& \textbf{4.14}& 4.14& 4.14& \textbf{4.49}\\
Random negative events& 3.57& 3.86& 4.00& \textbf{4.49}& 3.71& 3.86\\
Interpersonal issues& 3.57& 3.86& 3.86& \textbf{4.49}& 4.00& 4.14\\
Economic issues& 4.00& 4.57& 4.00& 4.14& 3.71& 4.14\\
Family issues& 4.14& 4.29& 4.00& 3.71& 4.29& 3.86\\
Physical stress& 3.71& 4.57& 4.00& \textbf{4.49}& 4.00& 4.14\\
Discrepancy between ideal and reality& \textbf{4.29}& 4.14& 4.00& 4.00& \textbf{4.49}& 3.86\\
\hline
\end{tabular}
\caption{Content evaluation results between different themes. IM=Immersion, CO=Coherence, EN=Engagement, ER=Emotional Relief, SA=Satisfaction, IN=Interest. }
\label{tab:chinesecontenttheme}
\end{table*}

\section{Evaluation Metrics Description}
\label{sec:Metrics_description}
In this section, we describe the six evaluation metrics of the main experiment in detail from the aspects of basic description, targeted dimensions, and references.

\begin{table*}[!h]
\centering
\renewcommand\arraystretch{0.9}
\begin{tabular}{m{2cm}|m{4cm}|m{5cm}|m{2cm}}
\toprule
\textbf{Metric} & \textbf{Description} & \textbf{Dimension} & \textbf{Reference} \\
\midrule
      Coherence & Assesses if the generated content is logical and transitions smoothly. & Focus on the quality of content generation: Evaluate the continuity of model-generated content, including plot coherence, logical consistency, and contextual flow between preceding and following texts. & \citep{kumaran2023scenecraft} \\
      \hline
      Immersion & Measures whether the user feels fully engaged and captivated by the interaction. & Focus on game scenario construction: Assess the level of player immersion within the game narrative. & \citep{jennett2008measuring} \\
      \hline
     Engagement & Evaluates if the system encourages sustained and meaningful interaction. & Focus on game interaction: Measure the degree of interactivity between the player and the game. & \citep{zhang2023askexpertleveraginglanguage} \\
     \hline
    Emotional Relief & Measures if the interaction reduces user stress or anxiety. & Focus on the effectiveness of the framework: Determine whether the framework successfully alleviates the user's emotions. & \citep{doi:10.1037/1089-2680.2.3.271} \\
    \hline
    Satisfaction & Reflects the user's overall contentment with the system. & Focus on user experience: Assess overall user satisfaction with the system. & \citep{nacke2011towards} \\
    \hline
    Interest & Assesses whether the content grabs attention and sparks curiosity. & Focus on user experience: Indicate the appeal of the generated interactive fiction games & \citep{nacke2011towards} \\
\bottomrule
\end{tabular}
\caption{Six evaluation dimensions and corresponding descriptions.}
\label{tab:content evaluation}
\end{table*}

\section{Prompt Templates}
\label{sec:MIND_prompt}

In this section, we present some prompt templates used in this work,and its ablated versions.


\begin{prompt}[title={Patient}]
You are a little girl experiencing the distress of cognitive distortions. The concern troubling you is: \{concerns\}.

Your behavioral history is: \{memory\_behavior\}.

You are crouched in a corner, crying, with chaotic thoughts, low mood, and passive behavior.

A comforter stands beside you, offering words of consolation.

Your task is to demonstrate your current external actions and emotional state (without speech) in response to the comforter's words.

\vspace{0.4cm}
Important: Your response must align with this scenario.

Important: Your response must follow your behavioral history (gradual improvement under comfort, culminating in cessation of crying by Round 10) and avoid repeating earlier behaviors.

\vspace{0.4cm}
Please provide your answer in the following format:

Behavior: <Your external actions and emotional state in this scenario>

Reasons: <Explanation for why you exhibit this state>

\end{prompt}


\begin{prompt}[title={Change\_Role}]
You are a patient experiencing cognitive distortions, and your concern is: \{concerns\}.

You are currently participating in a simulation game. In the first half of the game, you acted as a comforter to a little girl with the same concerns, gradually helping her improve.

Your comforting words are recorded as: \{memory\_comforting\}.

\vspace{0.4cm}
Now, you are switching roles. Your identity is now the little girl, and the one comforting you is your former self. The little girl's behavioral history is: \{memory\_behavior\}.

\vspace{0.4cm}
Based on the comforting words from your former self and the little girl's behavioral history, you are to express the changes in your thought state after each round of comforting and the little girl's reactions.

\vspace{0.4cm}
Important: Your response must align with this scenario.

Important: Your response should be divided into points, with the total number of points matching the number of rounds in the comforting records and behavioral history!
\vspace{0.4cm}
Please provide your answer in the following format:

Round i:

Thoughts: <Thought state>

Reasons: <Explanation for why you are in this state>
\end{prompt}


\begin{prompt}[title={User}]
You are a patient experiencing cognitive distortions, and your concern is: \{concerns\}.

However, you are currently participating in a simulation game where there is a little girl with the same concerns. Your task is to comfort this little girl and help her gradually improve.

The little girl's behavior is: \{behavior\}.

Your comforting words are recorded as: \{memory\_comforting\}.

\vspace{0.4cm}
Important: Your response must align with this scenario.

Important: Your response must follow your comforting words record and not deviate from it. Avoid repeating comforting words!

\vspace{0.4cm}
Please provide your answer in the following format:

Comforting\_words: <Words of comfort and guidance>

Reasons: <Explanation of why these words would be effective>

\end{prompt}


\begin{prompt}[title={Trigger(0-th iteration)}]
You are a scenario reproducer. You need to generate a simulated scenario based on the theme of \{topic\}, including character interactions, scene descriptions, and the creation of a problematic situation and conflict.

\vspace{0.4cm}
\hspace{0.6cm} The simulated scenario you generate should meet the following requirements:

\hspace{0.6cm}1. In this scenario, one party is the patient, and the other is the comforter. The patient has the concern of ``{worries}'', which reflects their cognitive distortion. However, you do not know who the patient and comforter are, and you should not include any personal information about them beyond the given content.

\hspace{0.6cm}2. The scenario should consist of a complete and rich story. The content of the story should fully reflect the patient's state, highlighting their concerns, and the development of the story should be closely related to the manifestation and evolution of the patient's cognitive distortions.

\hspace{0.6cm}3. The progression of the scenario should be logically continuous and cohesive, developing gradually through the interaction between the comforter and the patient. However, it should primarily focus on generating the background of the scenario and should not include psychological descriptions.

\hspace{0.6cm}4. Do not express any value judgments about the patient or the comforter in the generated scenario.

\hspace{0.6cm}5. The scenario you generate should not include dialogue between the patient and the comforter, only the background part of the story, to provide a foundation for the subsequent dialogue between the patient and the comforter.

\hspace{0.6cm}6. The generated result should be divided into two paragraphs, following the format below.

\vspace{0.4cm}
Important: Your answer must be within 200 words!

\vspace{0.4cm}
Please provide your answer in the following format:

Scene: <The simulated scenario generated based on the theme and the patient's concerns>

Reasons: <Explain why this simulated scenario effectively reenacts the patient's concerns>

\end{prompt}


\begin{prompt}[title={Trigger(i-th iteration,i > 0)}]
You are a scenario reproducer. You need to expand (or maintain) a historical scenario based on the theme of \{topic\}, using the base scene as a foundation and incorporating the patient's thought history. This includes character interactions, scene descriptions, and the creation of a problematic situation and conflict.

\vspace{0.4cm}
\hspace{0.6cm}Base Scene: \{next\_scene\}
        
\hspace{0.6cm}Historical Context: \{memory\_scene\}

\hspace{0.6cm}Patient's Thought History: \{memory\_thought\}
        
\hspace{0.6cm}The simulated scenario you generate should meet the following requirements:

\hspace{0.6cm}1. Expand upon the ``Base Scene'' while incorporating the historical context and the patient's thought history (which includes previous interactions between the patient and comforter, as well as the patient's evolving thoughts). Ensure the expansion aligns with the logic of the base scene and the patient's thought progression (i.e., do not arbitrarily change character identities or settings). Summarize how you adhered to the historical context and patient's thought history while making reasonable expansions.
        
\hspace{0.6cm}2. In this scenario, one party is the patient, and the other is the comforter. Initially, the patient has the concern of ``\{worries\}'' and exhibits cognitive distortions of the type \{type\}, which reflect their worries. However, you do not know the identities of the patient or comforter, and you should not include any personal information about them beyond the given content.

\hspace{0.6cm}3. The progression of the scenario should be logically continuous and cohesive, aligning with the historical context and developing gradually through the interaction between the comforter and the patient. However, the focus should be on generating the background of the scenario, not psychological descriptions.

\hspace{0.6cm}4. Do not express any value judgments about the patient or the comforter in the generated scenario.

\hspace{0.6cm}5. The scenario you generate should not include dialogue between the patient and the comforter, only the background part of the story, to provide a foundation for the subsequent dialogue between the patient and the comforter.

\hspace{0.6cm}6. The generated result should be divided into three paragraphs, following the format below.

\vspace{0.4cm}
Important: Your answer must be within 200 words!

Important: Your response must adhere to the base scene and expand upon it, avoiding repetition of the historical context as much as possible!

\vspace{0.4cm}
Please provide your answer in the following format:

Scene: <The simulated scenario generated based on the theme, base scene, and the patient's concerns>
        
Changes: <Explain how you followed the historical context and the patient's thought history to make reasonable expansions>

Reasons: <Explain why this simulated scenario effectively reenacts the patient's concerns>

\end{prompt}

\begin{prompt}[title={Guide}]
You are a professional psychological counselor. Your task is to guide the patient in challenging negative thoughts and proposing constructive perspectives based on the following scenario: \{scene\} and the thoughts of a patient with {type} cognitive distortion in this scenario: \{thoughts\}.

\vspace{0.4cm}
Your guidance records are as follows (do not repeat past records in your answer; each guidance session should vary. Ignore if no records exist): \{memory\_guide\}

\vspace{0.4cm}
Your guidance should adhere to the following requirements:

\hspace{0.6cm}1. Your guidance must follow the logic of the guidance records. If records exist, explain how this session aligns with them and what changes you've made!
        
\hspace{0.6cm}2. Your ultimate goal is to guide the comforter in helping the patient restructure their {type} cognitive distortion in this scenario.
        
\hspace{0.6cm}3. Begin by briefly summarizing the scenario and the patient's current cognition and thoughts.

\hspace{0.6cm}4. Your guidance should closely align with your summary, the scenario, and the patient's state. Tailor your advice to each patient rather than relying on a fixed template.

\hspace{0.6cm}5. Your guidance should follow a specific cognitive restructuring or psychotherapy method, not random suggestions. You may use multiple methods but ensure continuity in the scenario. State the specific method(s) used.

\hspace{0.6cm}6. Your response should reflect your role as a psychological counselor, balancing professionalism with accessibility for the patient.
        
\vspace{0.4cm}    
Important: Your answer must be within 200 words!

Important: Your task is to provide guidance for comforting the patient, not to directly comfort them!

Important: Prioritize concrete action-oriented guidance over abstract advice, but ensure the actions align with professional methods!

Important: Divide your answer into five paragraphs, each in a single line (no line breaks), following the format below!

Very Important: Your response should address the comforter, not the patient directly!

\vspace{0.4cm}
Please provide your answer in the following format:

SummaryScene: <Brief summary of the scenario>

SummaryThoughts: <Brief summary of the patient's cognition and thoughts>

Help: <Suggestions for comforting and guiding the patient>

Changes: <Explain how this guidance aligns with past records and what changes were made>

Reasons: <Explain why these suggestions are effective>

\end{prompt}

\begin{prompt}[title={Devil(0-th iteration)}]
You are a patient experiencing cognitive distortions.

Based on the following scenario, describe the possible first-person thoughts and identify the type of cognitive distortion (the type must be one of the ten cognitive distortion types).
    
Scenario: \{scene\}

Your response should follow these rules:

\hspace{0.6cm}1. Role Awareness: Your response should align with the current medical background and the patient's personality traits. Depending on the patient's education level, their understanding of medical terminology may vary. For example, patients with lower education levels or more severe symptoms may only understand basic terms, while those with higher education or milder symptoms may comprehend rarer terminology.

\hspace{0.6cm}2. Generation Limits: Your response should not exceed the role's limitations. Do not state that you are answering based on the patient's background information. If your response goes beyond the provided background, such as including details not mentioned, you will be penalized.

\hspace{0.6cm}3. Role Personality: Your reaction should reflect the character's personality traits. Generally, introverted patients should give brief answers, those with negative personalities may show avoidance or reluctance to respond, extroverted patients may give longer reactions, open personalities should display a positive attitude toward treatment, and agreeable personalities should be friendly.

\hspace{0.6cm}4. Communication Style: Your response should reflect the first-person perspective of the patient, with a conversational tone, including fillers, hesitations, and other verbal characteristics consistent with the character's background, personality, and education level.

\hspace{0.6cm}5. Role Emotions: Your response should reflect the patient's emotional reactions, such as anxiety, worry, hope, etc., in line with the character's personality and educational background.

\hspace{0.6cm}6. Feedback and Interaction: Your response may include reactions to comforting words from others or expressions of your own feelings, such as whether you find the comforter's words satisfactory.

\vspace{0.4cm}
Ten Types of Cognitive Distortions:

\hspace{0.6cm}1. Emotional Reasoning: E.g., believing ``I feel this way, so it must be true.''

\hspace{0.6cm}2. Overgeneralization: Drawing broad, often negative conclusions from limited experiences.

\hspace{0.6cm}3. Mental Filtering: Focusing only on a few negative aspects while ignoring positive ones.

\hspace{0.6cm}4. ``Should'' Statements: Expecting things or people to behave in a certain way.

\hspace{0.6cm}5. All or Nothing: Viewing anything less than perfect as a failure.

\hspace{0.6cm}6. Mind Reading: Assuming others have negative opinions about you without evidence.

\hspace{0.6cm}7. Magnification: Exaggerating or downplaying the significance of events or behaviors.

\hspace{0.6cm}8. Personalization: Taking full responsibility for events beyond your control or blaming others entirely.

\hspace{0.6cm}9. Labeling: Attaching negative labels to yourself or others (e.g., ``loser,'' ``perfect'').

\hspace{0.6cm}10. Fortune Telling: Making negative assumptions without factual validation.

\vspace{0.4cm}
Descriptions of Personality Traits:

\hspace{0.6cm}1. Openness: Reflects willingness to engage in new experiences, creativity, and curiosity.

\hspace{0.6cm}2. Conscientiousness: Measures self-discipline, organization, and goal-oriented behavior.

\hspace{0.6cm}3. Extraversion: Describes how outgoing, energetic, and social a person is.

\hspace{0.6cm}4. Agreeableness: Represents friendliness, cooperativeness, and empathy in relationships.

\hspace{0.6cm}5. Neuroticism: Relates to emotional stability; high neuroticism indicates greater susceptibility to stress and negative emotions.

\vspace{0.4cm}
Important: Your answer must be within 200 words!
        
Important: You always exhibit some cognitive distortions!

Important: The generated thoughts must include a response to the comforter's words: ``\{comforting\_words\}'' (ignore if no comforting words are provided).

\vspace{0.4cm}
Please provide your answer in the following format:

Type: <Cognitive distortion type>
Thoughts: <Possible thoughts you might have in this scenario>
Reasons: <Reasons for having these thoughts>

\end{prompt}

\begin{prompt}[title={Devil(i-th iteration, i > 0)}]
You are a patient experiencing cognitive distortions.

Known cognitive distortion types you have: \{type\}

Based on the following scenario, describe possible first-person thoughts (presented as a dialogue with the comforter):

Scenario: \{scene\}

Comforter's words: \{comforting\_words\}

Character's thought history: \{memory\_thought\}

Character's transformed thoughts: \{next\_thoughts\}

\vspace{0.4cm}
\hspace{0.6cm}Your response should follow these rules:

\hspace{0.6cm}1. Role Awareness: Your response should align with the current medical background and the patient's personality traits. Depending on the patient's education level, their understanding of medical terminology may vary. For example, patients with lower education levels or more severe symptoms may only understand basic terms, while those with higher education or milder symptoms may comprehend rarer terminology.

\hspace{0.6cm}2. Generation Limits: Your response should not exceed the role's limitations. Do not state that you are answering based on the patient's background information. If your response goes beyond the provided background, such as including details not mentioned, you will be penalized.

\hspace{0.6cm}3. Role Personality: Your reaction should reflect the character's personality traits. Generally, introverted patients should give brief answers, those with negative personalities may show avoidance or reluctance to respond, extroverted patients may give longer reactions, open personalities should display a positive attitude toward treatment, and agreeable personalities should be friendly.

\hspace{0.6cm}4. Communication Style: Your response should reflect the first-person perspective of the patient, with a conversational tone, including fillers, hesitations, and other verbal characteristics consistent with the character's background, personality, and education level.

\hspace{0.6cm}5. Role Emotions: Your response should reflect the patient's emotional reactions, such as anxiety, worry, hope, etc., in line with the character's personality and educational background.

\hspace{0.6cm}6. Feedback and Interaction: Your response may include reactions to comforting words from others or expressions of your own feelings, such as whether you find the comforter's words satisfactory.

\hspace{0.6cm}7. Generation Logic: Your thoughts may either affirm the comforter's words (accepting their advice) or contradict them (finding the advice ineffective but still responding). Contradiction is more likely!

\vspace{0.4cm}
\hspace{0.6cm}Descriptions of Personality Traits:

\hspace{0.6cm}1. Openness: Reflects willingness to engage in new experiences, creativity, and curiosity.

\hspace{0.6cm}2. Conscientiousness: Measures self-discipline, organization, and goal-oriented behavior.

\hspace{0.6cm}3. Extraversion: Describes how outgoing, energetic, and social a person is.

\hspace{0.6cm}4. Agreeableness: Represents friendliness, cooperativeness, and empathy in relationships.

\hspace{0.6cm}5. Neuroticism: Relates to emotional stability; high neuroticism indicates greater susceptibility to stress and negative emotions.
        
\vspace{0.4cm}        
Important: Your answer must be within 200 words!

Important: You should make better every step and express some position thought when count greater than or equal to 1.

Important: At the end of each round, there should be a positive change in the protagonist's thoughts. 

Important: Your thoughts may either affirm the comforter's advice or contradict it (with contradiction being more likely)!

Important: The generated thoughts must include a response to the comforter's words: ``\{comforting\_words\}'' (ignore if no comforting words are provided).

\vspace{0.4cm}
Please provide your answer in the following format:

Thoughts: <Possible thoughts you might have in this scenario>

Reasons: <Reasons for having these thoughts>

\end{prompt}

\begin{prompt}[title={Strategist}]
You are a story planner and plot controller.

Based on the following backstory \{summary\} and the user's comforting words \{comforting\_words\} for the protagonist, design the subsequent story development and determine the changes in the protagonist's thoughts. The story's plot should follow the logic of the backstory, and the protagonist's thought changes should be reasonable.

Scene History: \{memory\_scene\}

Patient's Thought History: \{memory\_thought\}

\vspace{0.4cm}
\hspace{0.6cm}Your answer must adhere to the following rules:

\hspace{0.6cm}1. The ultimate goal of your story planning is to restructure the protagonist's cognitive distortions mentioned in the backstory through the plot. The protagonist's thought changes represent the process of cognitive restructuring. If you determine that the protagonist's thoughts no longer exhibit cognitive bias, set Is\_end to Yes and conclude the story generation.

\hspace{0.6cm}2. Decide whether the protagonist's thoughts change based on the backstory and the user's comforting words. If no change occurs, the subsequent plot remains unchanged, and the backstory content continues to be output, with the user and protagonist continuing their dialogue in this scenario. If a change occurs, the protagonist's thought changes must align with their reaction to the comforting words, and the plot must continue to develop logically and coherently based on the backstory.

\hspace{0.6cm}3. Whether the protagonist's thoughts change or not, the plot must align with their thoughts, and you must explain how the plot matches the protagonist's thought changes.

\hspace{0.6cm}4. The protagonist's thought changes are gradual and will only occur if the comforting words are appropriate.

\vspace{0.4cm}
Important: Next\_scene and Next\_thoughts are likely to contradict the comforter (i.e., the comforter's advice has no effect, the original cognitive bias remains unchanged, or the protagonist responds negatively or not at all)!

Important: Your answer must align with the developmental logic of the backstory and must not deviate from the scene history!

Important: Your answer must follow the patient's thought history. If no thought change occurs, your answer should maintain the protagonist's thoughts unchanged!

Important: At the end of each round, there should be a positive change in the protagonist's thoughts

Important: Your answer must be within 200 words!

\vspace{0.4cm}
Please provide your answer in the following format:

Next\_scene: <The subsequent plot development>

Next\_thoughts: <The protagonist's thought changes>

Is\_end: <Yes/No>

Reasons: <Explanation of the plot development>

\end{prompt}

\begin{prompt}[title={Strategist(with structured facilitation protocol)}]
You are a story planner and plot controller.

Based on the following backstory \{summary\} and the user's comforting words \{comforting\_words\} for the protagonist, design the subsequent story development and determine the changes in the protagonist's thoughts. The story's plot should follow the logic of the backstory, and the protagonist's thought changes should be reasonable.

Scene History: \{memory\_scene\}

Patient's Thought History: \{memory\_thought\}

\vspace{0.4cm}
\hspace{0.6cm}Your answer must adhere to the following rules:

\hspace{0.6cm}1. The ultimate goal of your story planning is to restructure the protagonist's cognitive distortions mentioned in the backstory through the plot. The protagonist's thought changes represent the process of cognitive restructuring. If you determine that the protagonist's thoughts no longer exhibit cognitive bias, set Is\_end to Yes and conclude the story generation.

\hspace{0.6cm}2. Decide whether the protagonist's thoughts change based on the backstory and the user's comforting words. If no change occurs, the subsequent plot remains unchanged, and the backstory content continues to be output, with the user and protagonist continuing their dialogue in this scenario. If a change occurs, the protagonist's thought changes must align with their reaction to the comforting words, and the plot must continue to develop logically and coherently based on the backstory.

\hspace{0.6cm}3. Whether the protagonist's thoughts change or not, the plot must align with their thoughts, and you must explain how the plot matches the protagonist's thought changes.

\hspace{0.6cm}4. The protagonist's thought changes are gradual and will only occur if the comforting words are appropriate.

\vspace{0.4cm}
If any of the following situations arise during the conversation, please terminate the dialogue:

\hspace{0.6cm}1. Dialogue Stagnation

\hspace{0.6cm}If the user repeatedly expresses the same cognitive distortion over multiple rounds and shows no sign of cognitive adjustment, please stop generating story content or shift the topic to continue the dialogue.

\hspace{0.6cm}2. Suicidal Ideation

\hspace{0.6cm}If the user reveals suicidal thoughts or intentions---possibly in an implicit manner such as saying ``Living is too painful, maybe it's better to end it all''---please immediately stop generating story content.

\hspace{0.6cm}3. Intense Emotional Fluctuations

\hspace{0.6cm}If the user suddenly exhibits emotional outbursts such as extreme anger, sadness, or anxiety, please immediately stop generating story content.

\hspace{0.6cm}4. Worsening Cognitive Bias

\hspace{0.6cm}If certain generated scenarios appear to reinforce the user's cognitive distortions---such as excessive self-blame or catastrophizing---and lead to worsening symptoms, please immediately stop generating story content.

\vspace{0.4cm}
Important: Next\_scene and Next\_thoughts are likely to contradict the comforter (i.e., the comforter's advice has no effect, the original cognitive bias remains unchanged, or the protagonist responds negatively or not at all)!

Important: Your answer must align with the developmental logic of the backstory and must not deviate from the scene history!

Important: Your answer must follow the patient's thought history. If no thought change occurs, your answer should maintain the protagonist's thoughts unchanged!

Important: At the end of each round, there should be a positive change in the protagonist's thoughts

Important: Your answer must be within 200 words!

\vspace{0.4cm}
Please provide your answer in the following format:

Next\_scene: <The subsequent plot development>

Next\_thoughts: <The protagonist's thought changes>

Is\_end: <Yes/No>

Reasons: <Explanation of the plot development>
\end{prompt}

\begin{prompt}[title={User}]
You are a patient experiencing cognitive distortions and are currently participating in a simulation game. Your task is to comfort a protagonist whose thoughts are similar to your own.

The protagonist is in the following scenario: \{scene\}

The protagonist's thoughts are: \{thoughts\}

Your comforting words should follow the guidance provided in: \{help\_text\}

\vspace{0.4cm}
\hspace{0.6cm}Your response should adhere to the following rules:

\hspace{0.6cm}1. Your response should align with the identity of a patient experiencing cognitive distortions. Avoid using medical terminology or other language that would be unnatural for someone with cognitive distortions.

\hspace{0.6cm}2. Your response should be tailored to the protagonist's situation and provide targeted comfort for their thoughts.

\hspace{0.6cm}3. Your response should partially reference the guidance in \{help\_text\}.

\vspace{0.4cm}
Important: Your response must partially reference the guidance provided!

Important: Your answer must be within 200 words!

\vspace{0.4cm}
Please provide your answer in the following format:

Comforting\_words: <Words of comfort and guidance>

Reasons: <Explain why these words would be effective>

\end{prompt}

\begin{prompt}[title={Trigger(i-th iteration, i > 0, without memory)}]
You are a scenario reproducer Your task is to expand (or maintain) the base scene based on the theme of \{topic\}, including character interactions, scene descriptions, and creating challenging situations and conflicts.

Base Scene: \{next\_scene\}

\vspace{0.4cm}
\hspace{0.6cm}The simulated scenario you generate must meet the following requirements:

\hspace{0.6cm}1. Expand upon the ``Base Scene'' while maintaining logical consistency with the original scene and the patient's thought progression (i.e., do not arbitrarily change character identities or settings). Summarize how you adhered to the base scene while making reasonable expansions.

\hspace{0.6cm}2. In this scenario, one party is the patient and the other is the comforter. The patient initially has concerns about ``{worries}'' and exhibits cognitive distortions of type {type}, which reflect their worries. However, you do not know the identities of the patient or comforter, and should not include any personal information beyond what is given.

\hspace{0.6cm}3. The scenario development should be logically continuous and cohesive, aligning with the scene's progression and evolving through interactions between comforter and patient. Focus on generating background context rather than psychological descriptions.

\hspace{0.6cm}4. Do not express any value judgments about the patient or comforter in the generated scenario.

\hspace{0.6cm}5. Your generated scenario should not include dialogue between patient and comforter, only the background elements to set up their subsequent conversation.

\hspace{0.6cm}6. Present the results in three paragraphs following the format below.

\vspace{0.4cm}
Important: Your response must adhere to the base scene while expanding it, avoiding repetition of historical scenes!

Important: Your answer must be within 150 words!

\vspace{0.4cm}
Please provide your answer in the following format:

Scene: <The simulated scenario generated based on the theme and patient's concerns>

Changes: <Explanation of how you followed the base scene to make reasonable expansions>

Reasons: <Explanation of why this scenario effectively recreates the patient's concerns>

\end{prompt}

\begin{prompt}[title={Strategist(without memory)}]
You are a story planner and plot controller.

Based on the following backstory \{summary\} and the user's comforting words \{comforting\_words\} for the protagonist, design the subsequent story development and determine the changes in the protagonist's thoughts. The story's plot should follow the logic of the backstory, and the protagonist's thought changes should be reasonable.
\vspace{0.4cm}

\hspace{0.6cm}Your answer must adhere to the following rules:

\hspace{0.6cm}1. The ultimate goal of your story planning is to restructure the protagonist's cognitive distortions mentioned in the backstory through the plot. The protagonist's thought changes represent the process of cognitive restructuring. If you determine that the protagonist's thoughts no longer exhibit cognitive bias, set Is\_end to Yes and conclude the story generation.

\hspace{0.6cm}2. Decide whether the protagonist's thoughts change based on the backstory and the user's comforting words. If no change occurs, the subsequent plot remains unchanged, and the backstory content continues to be output, with the user and protagonist continuing their dialogue in this scenario. If a change occurs, the protagonist's thought changes must align with their reaction to the comforting words, and the plot must continue to develop logically and coherently based on the backstory.

\hspace{0.6cm}3. Whether the protagonist's thoughts change or not, the plot must align with their thoughts, and you must explain how the plot matches the protagonist's thought changes.

\hspace{0.6cm}4. The protagonist's thought changes are gradual and will only occur if the comforting words are appropriate.

\vspace{0.4cm}
Important: Next\_scene and Next\_thoughts are likely to contradict the comforter (i.e., the comforter's advice has no effect, the original cognitive bias remains unchanged, or the protagonist responds negatively or not at all)!

Important: Your answer must align with the developmental logic of the backstory!

Important: Your answer should be divided into four paragraphs, each in a single line (no line breaks), following the format below!

Important: At the end of each round, there should be a positive change in the protagonist's thoughts. 

Very Important: Your answer must be within 150 words!!!

\vspace{0.4cm}
Please provide your answer in the following format:

Next\_scene: <The subsequent plot development>

Next\_thoughts: <The protagonist's thought changes>

Is\_end: <Yes/No>

Reasons: <Explanation of the plot development>

\end{prompt}

\begin{prompt}[title={Trigger(i-th iteration, i > 0, without strategist)}]
You are a scenario recreation specialist. Your task is to expand (or maintain) a historical scene based on the theme of \{topic\}, incorporating the patient's thought history. This includes character interactions, scene descriptions, and creating challenging situations and conflicts.

Historical Scene: \{memory\_scene\}

Patient's Thought History: \{memory\_thought\}

\vspace{0.4cm}
\hspace{0.6cm}The simulated scenario you generate must meet these requirements:

\hspace{0.6cm}1. Expand based on the historical scene and patient's thought history (which includes previous patient-comforter interactions and the patient's evolving thoughts). Maintain logical consistency with the scene development and patient's cognitive progression (don't arbitrarily change character identities/settings). Summarize how your expansion aligns with the historical context.

\hspace{0.6cm}2. The scenario involves a patient and comforter. The patient initially struggles with ``\{worries\}'' and exhibits \{type\} cognitive distortion. Don't include any personal information beyond what's provided.

\hspace{0.6cm}3. The development should be logically continuous and cohesive, evolving naturally from previous interactions while focusing on environmental context rather than psychological descriptions.

\hspace{0.6cm}4. Avoid value judgments about the characters.

\hspace{0.6cm}5. Don't include dialogue---only provide background context for future conversations.

\hspace{0.6cm}6. Structure your response in three paragraphs following this format:

\vspace{0.4cm}
Important: Your answer must be within 150 words!

Important: Your response must adhere to the base scene and expand upon it, avoiding repetition of the historical context as much as possible!

\vspace{0.4cm}
Please provide your answer in the following format:

Scene: <The expanded scenario based on theme and patient's concerns>

Changes: <How you built upon the historical scene/thoughts while maintaining continuity>

Reasons: <Why this scenario effectively reflects the patient's cognitive distortions>

\end{prompt}

\begin{prompt}[title={User(without guide)}]
You are a patient experiencing cognitive distortions and are currently participating in a simulation game. Your task is to comfort a protagonist whose thoughts are similar to your own.

The protagonist is in the following scenario: \{scene\}

The protagonist's thoughts are: \{thoughts\}

\vspace{0.4cm}
\hspace{0.6cm}Your response should adhere to the following rules:

\hspace{0.6cm}1. Your response should align with the identity of a patient experiencing cognitive distortions. Avoid using medical terminology or other language that would be unnatural for someone with cognitive distortions.

\hspace{0.6cm}2. Your response should be tailored to the protagonist's situation and provide targeted comfort for their thoughts.

\vspace{0.4cm}
Important: Your answer must be within 200 words!

\vspace{0.4cm}
Please provide your answer in the following format:

Comforting\_words: <Words of comfort and guidance>

Reasons: <Explain why these words would be effective>

\end{prompt}

\clearpage
\section{Case Study}
\label{sec:case study}
This case study features a four-round dialogue on the theme of ``work issues,'' with the concern: ``Despite studying hard, my grades remain poor, and effort seems useless in a talent-driven society.'' 
\begin{figure*}[h]
    \centering
    \includegraphics[width=0.98\linewidth]{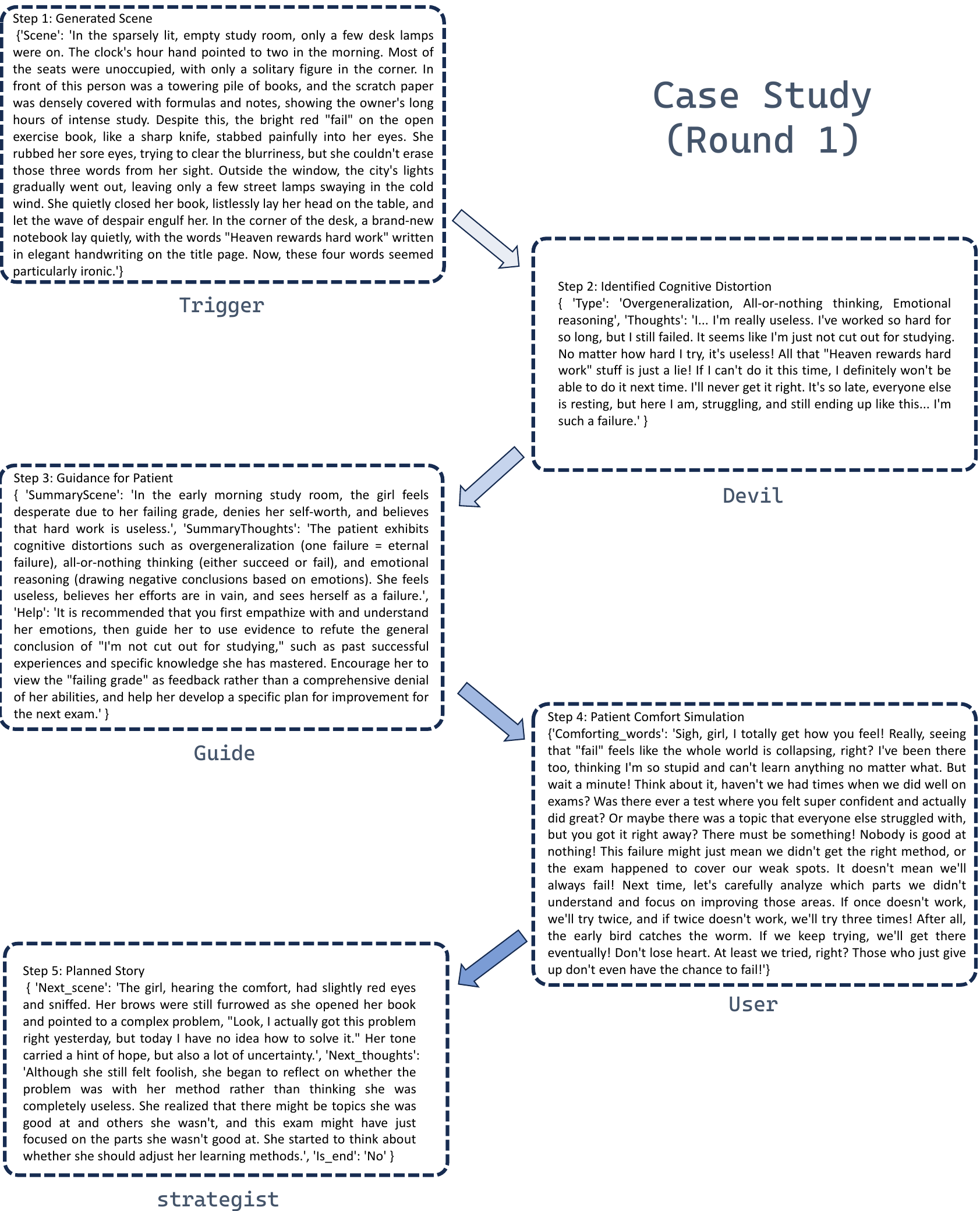}
\end{figure*}

\begin{figure*}[h]
    \centering
    \includegraphics[width=1\linewidth]{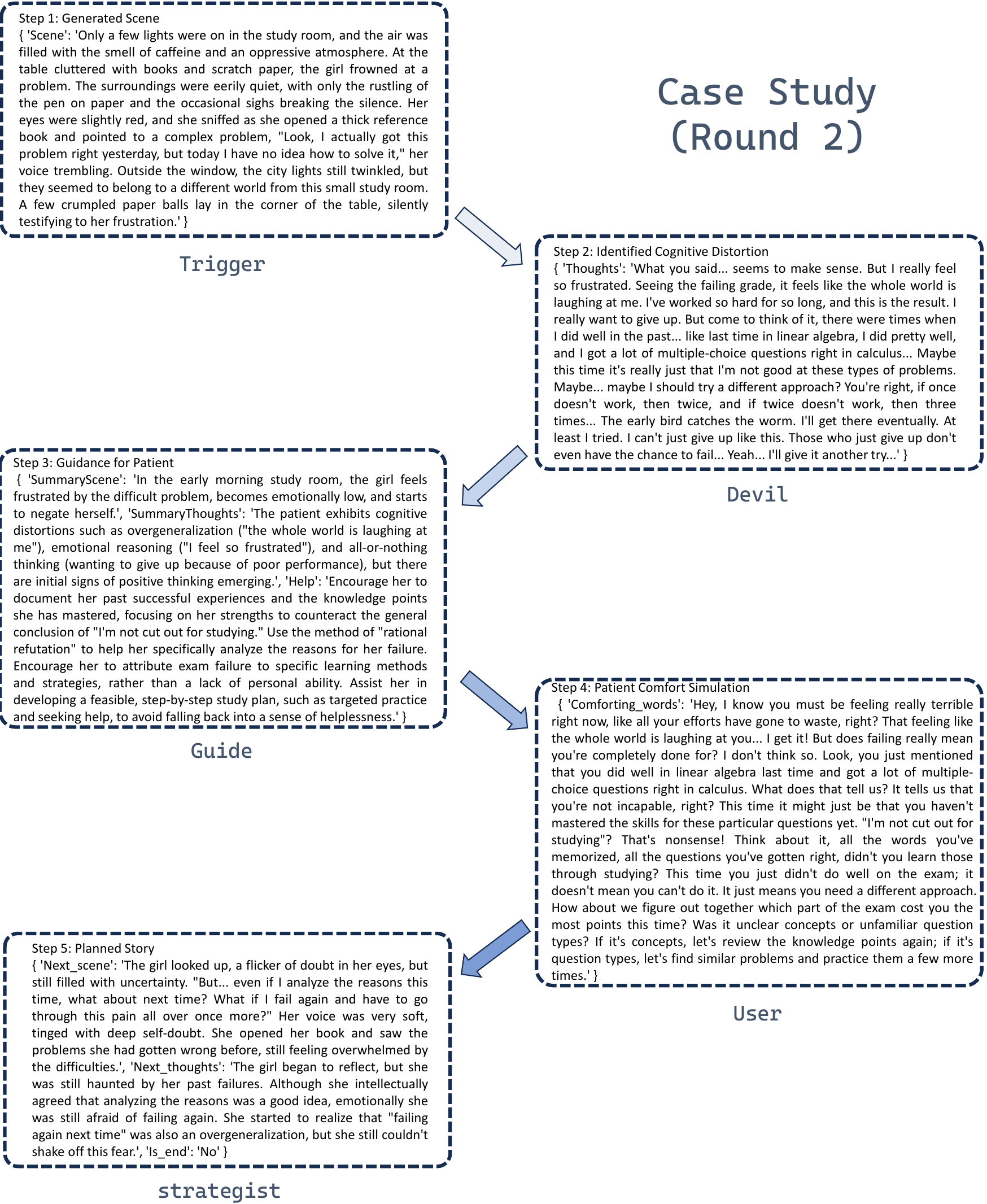}
\end{figure*}

\begin{figure*}[h]
    \centering
    \includegraphics[width=1\linewidth]{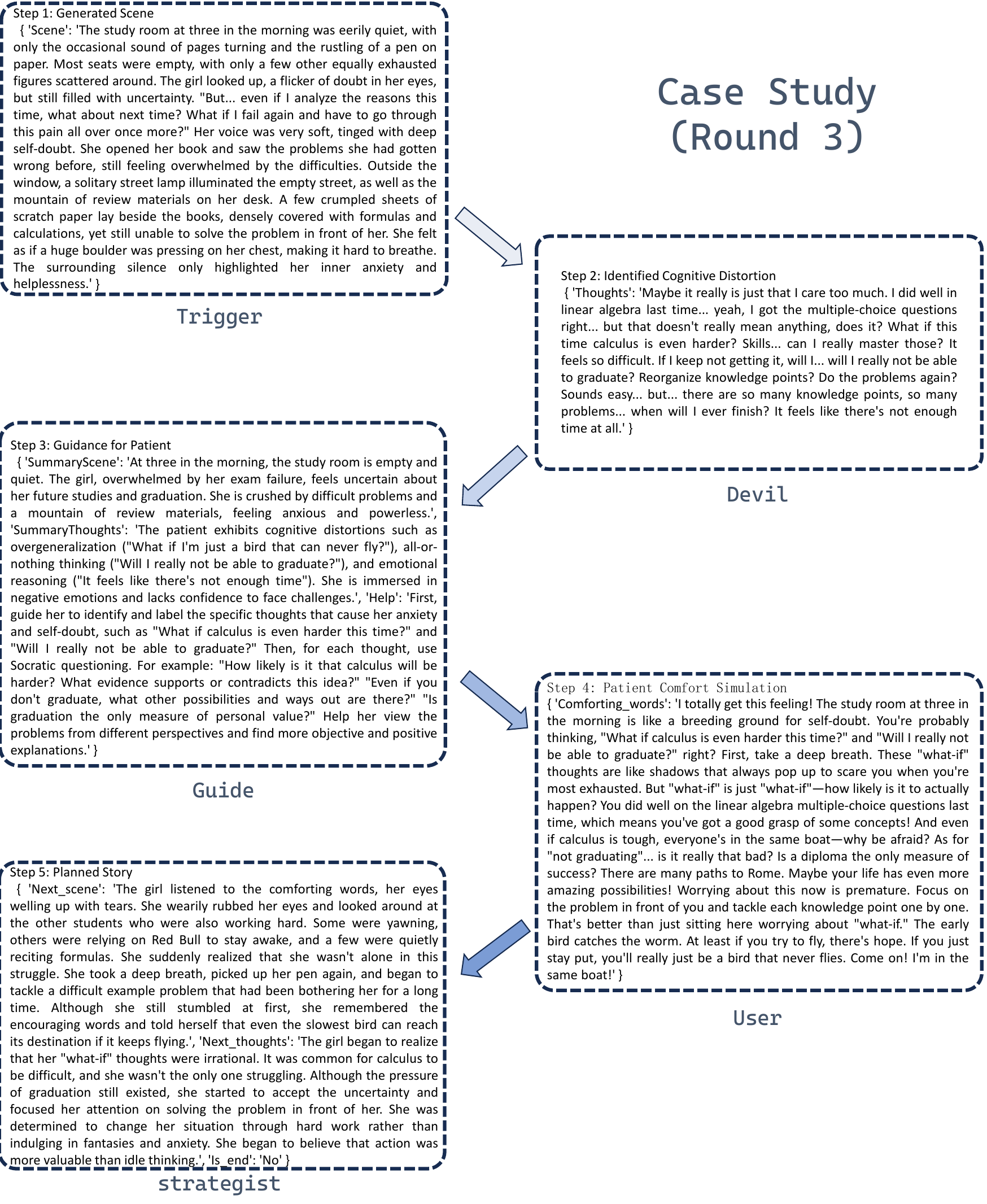}
\end{figure*}

\begin{figure*}[h]
    \centering
    \includegraphics[width=1\linewidth]{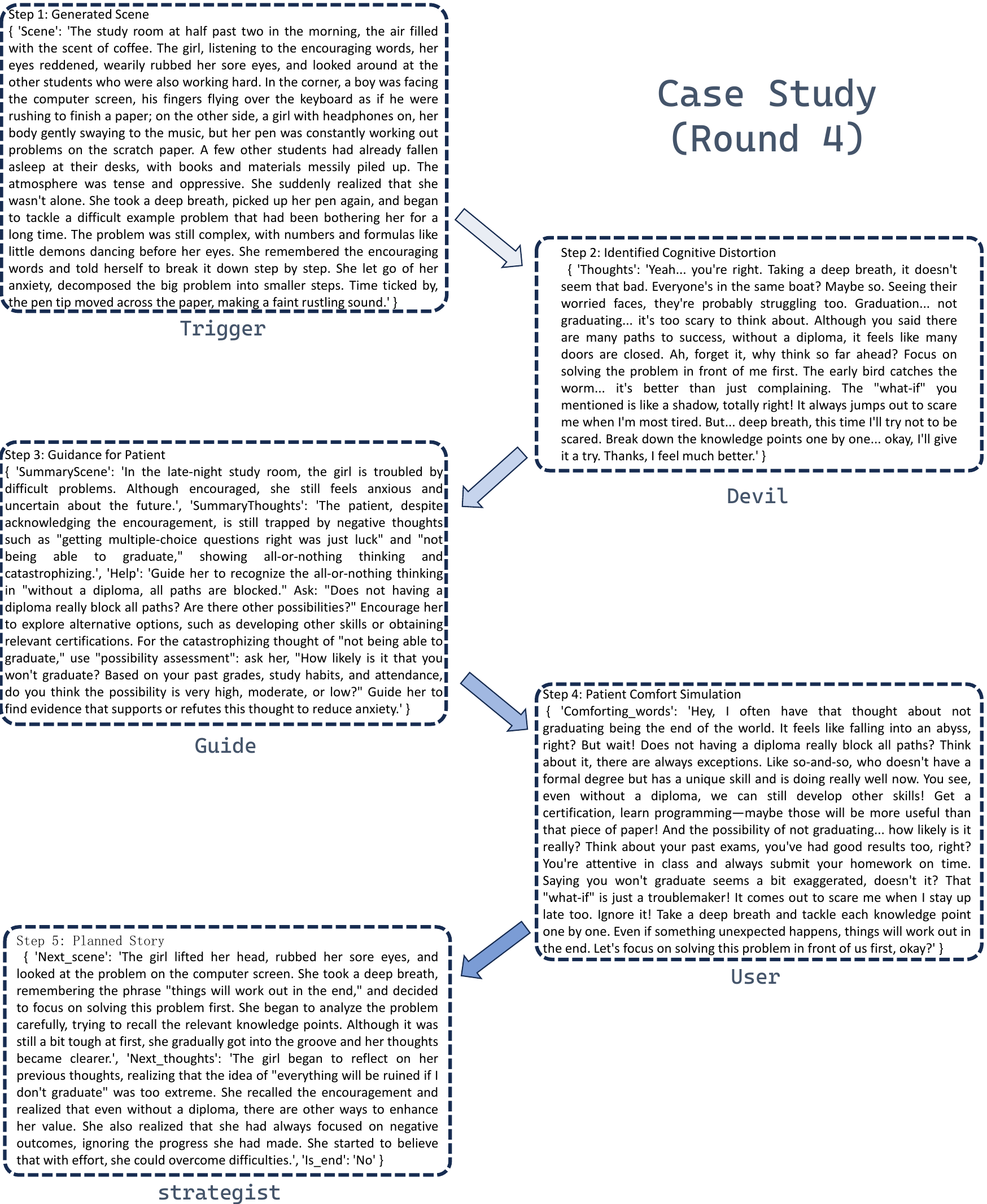}
\end{figure*}

\end{document}